\documentclass{article}

\usepackage[nonatbib, preprint]{neurips_2023}

\usepackage{multirow,tabularx,graphicx,array}

\usepackage{enumitem}

\usepackage{arydshln}

\usepackage[utf8]{inputenc} 
\usepackage[T1]{fontenc}    
\usepackage{hyperref}       
\usepackage{url}            
\usepackage{booktabs}       
\usepackage{amsfonts}       
\usepackage{nicefrac}       
\usepackage{microtype}      
\usepackage{xcolor}         

\usepackage{cite}
\usepackage{hyperref}
\usepackage{amsmath,amssymb,amsfonts}
\usepackage{graphicx}
\usepackage{textcomp}
\usepackage{comment}
\usepackage{xcolor}

\usepackage{balance}
\def\BibTeX{{\rm B\kern-.05em{\sc i\kern-.025em b}\kern-.08em
    T\kern-.1667em\lower.7ex\hbox{E}\kern-.125emX}}

\makeatletter
\let\NAT@parse\undefined
\makeatother

\usepackage{tabularx}
\usepackage{url}
\usepackage[utf8]{inputenc} 
\usepackage[T1]{fontenc}    
\usepackage{booktabs}       
\usepackage{amsfonts}       
\usepackage{nicefrac}       
\usepackage{microtype}      
\usepackage{xcolor}         
\usepackage{amsmath}
\usepackage{adjustbox}
\usepackage{amsmath,amssymb}
\usepackage{graphicx}
\usepackage[utf8]{inputenc}
\usepackage{amsfonts}
\usepackage{graphicx}
\usepackage{textcomp}

\usepackage{algorithm}
\usepackage[noend]{algpseudocode}
\newcommand{\algalign}[2]

\usepackage{cellspace}
\usepackage{graphicx}
\usepackage{caption}
\usepackage{subcaption}

\newcommand{\qedsymbol}{\hspace{\fill}\rule{1.5ex}{1.5ex}}
\definecolor{square_col}{rgb}{0.221197, 0.016497, 0.602083}
\definecolor{triangle_col}{rgb}{0.706178, 0.178437, 0.553657}
\definecolor{edge_col}{rgb}{0.961336, 0.548636, 0.275305}

\newcommand{\beq}{\begin{equation}}
\newcommand{\eeq}{\end{equation}}

\DeclareMathOperator*{\argmax}{arg\,max}
\title{From Latent Graph to Latent Topology Inference:\\ Differentiable Cell Complex Module}

\author{Claudio Battiloro$^{1,}$\thanks{Equal contribution. Corresponding authors, email: \{claudio.battiloro, indro.spinelli\}@uniroma1.it} \\
\and 
\textbf{Indro Spinelli}$^{1,*}$ \\
\and 
\textbf{Lev Telyatnikov}$^{1}$ \\
\and
\textbf{Michael Bronstein}$^2$ \\
\and
\textbf{Simone Scardapane}$^1$ \\
\and
\textbf{Paolo Di Lorenzo}$^1$\\
\and
$^1$ Sapienza University of Rome \quad $^2$ University of Oxford 
}

\begin{document}

\maketitle

\begin{abstract}
Latent Graph Inference (LGI) relaxed the reliance of Graph Neural Networks (GNNs) on a given graph topology by dynamically learning it. However, most of LGI methods assume to have a (noisy, incomplete, improvable, ...) input graph to rewire and can solely learn regular graph topologies. In the wake of the success of  Topological Deep Learning (TDL), we study Latent Topology Inference (LTI) for learning higher-order cell complexes (with sparse and not regular topology) describing multi-way interactions between data points. To this aim, we introduce the Differentiable Cell Complex Module (DCM), a novel learnable function that computes cell probabilities in the complex to improve the downstream task. We show how to integrate DCM with cell complex message passing networks layers and train it in a end-to-end fashion, thanks to a two-step inference procedure that avoids an exhaustive search across all possible cells in the input, thus maintaining scalability. Our model is tested on several homophilic and heterophilic graph datasets and it is shown to outperform other state-of-the-art techniques, offering significant improvements especially in  cases where an input graph is not provided.
\end{abstract}

\section{Introduction}

Graph Neural Networks (GNNs) are a versatile tool exploited in a wide range of fields, such as computational chemistry ~\cite{gilmer2017neural}, physics simulations \cite{shlomi2021gnnphysics}, and social networks \cite{xia2021socialgnn}, just to name a few. GNNs have shown remarkable performance in learning tasks where data are represented over a graph domain, due to their ability to combine the flexibility of neural networks with prior knowledge about data relationships, expressed in terms of the underlying graph topology. The literature on GNNs is extensive and encompasses various techniques, typically categorized into spectral ~\cite{Bruna19} and non-spectral ~\cite{gori2005new} methods. The basic idea behind GNNs is to learn node (and/or) edge attributes representations using local aggregation based on the graph topology, i.e. message passing networks in their most general form \cite{gilmer2017mp}. By leveraging this feature, GNNs have achieved outstanding results in several tasks, including node and graph classification ~\cite{kipf2016semi}, link prediction ~\cite{zhang2018link}, and more specialized tasks such as protein folding ~\cite{jumper2021highly} and neural algorithmic reasoning ~\cite{velivckovic2021neural}. 

The majority of GNNs assume the graph topology to be fixed (and optimal) for the task at hand, therefore the focus is usually on designing more sophisticated architectures with the aim of improving the message passing process. Very recently, a series of works \cite{kazi2022dgm,borde2023latent,topping2022understanding} started to investigate techniques for Latent Graph Inference (LGI), where the intuition is that data can have some underlying but unknown (latent) graph structure, mainly in cases where only a point cloud of data is available but also when the given graph is suboptimal for the downstream task. At the same time, the field of Topological Deep Learning (TDL) \cite{barbarossa2020topological,hajij2023topological} started to gain interest, motivated by the fact that many systems are characterized by higher-order interactions  that cannot be captured by the intrinsically pairwise structure of graphs. Topological Neural Networks exploit tools and objects from (algebraic) topology to encode these multi-way relationships, e.g. simplicial ~\cite{bodnar2021weisfeiler,giusti22}, cell \cite{bodnarcwnet}, or combinatorial complexes \cite{hajij2023topological}. However, TDL techniques usually incorporate graphs in higher-order complexes by means of deterministic lifting maps, assigning higher-order cells to cliques or induced cycles of the graph, thus implicitly assuming that these (task-agnostic, deterministic) complexes are optimal for the downstream task. In addition, whenever an input graph is not available, these strategies scale poorly in the size of the input set by requiring an exhaustive (combinatorial) search over all possible cells, making them unfeasible for even medium-sized sets of data-points.

\textbf{Contribution.} In this paper, we introduce the concept of {\em Latent Topology Inference} (LTI) by generalizing LGI to higher-order  complexes. The goal of LTI is not (only) learning a graph structure describing  pairwise interactions  but rather learning a higher-order complex describing multi-way interactions among data points. As a first instance of LTI, we introduce the {\bf Differentiable Cell Complex Module (DCM)}, a novel deep learning architecture that dynamically learns a cell complex to improve the downstream task. The DCM implements a two-step inference procedure to alleviate the computational burden: first, learning the $1$-skeleton of the complex (i.e., a graph) via a novel improved version of the Differentiable Graph Module (DGM) \cite{kazi2022dgm}, and then  learning which higher-order cells (polygons) should be included in the complex. Both steps leverage message passing (at node and edge levels) and a sparse sampling technique based on the $\alpha$-entmax class of functions, which allows overcoming the limitation of the original DGM, capable of learning only regular graph topologies. We generalize the training procedure of the DGM \cite{kazi2022dgm} to train the DCM in a end-to-end fashion. The DCM is tested on several homophilic and heterophilic datasets and it is shown to outperform other state-of-the-art techniques, offering significant improvements 
{\em both when the input graph is provided or not.} 
In particular, accuracy gains on heterophilic benchmarks with provided input graphs indicate that the DCM leads to robust performance even when the input graph does not fit the data well. 

\begin{figure}[t]
  \centering
\includegraphics[width=1.00\columnwidth]{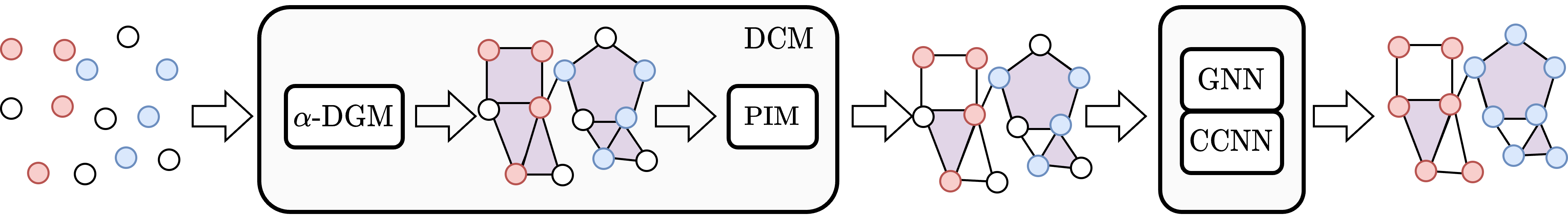}
  \caption{The proposed two-step procedure for Latent Topology Inference (LTI) via regular cell complexes. The Differentiable Cell Complex Module (DCM) is a function that first learns a graph describing the pairwise interactions among data points via the $\alpha$-Differentiable Graph Module ($\alpha$-DGM), and then it leverages the graph  as the 1-skeleton  of a regular cell complex whose 2-cells (polygons), describing multi-way interactions among data points, are learned via the Polygon Inference Module (PIM). The inferred topology is then used in two message passing networks, at node (Graph Neural Network, GNN) and edge (Cell Complex Neural Network, CCNN) levels to solve the downstream task. The whole architecture is trained in a end-to-end fashion.} 
  \label{fig:dcm_easy}
\end{figure}
\subsection{Related Works}
\paragraph{Latent graph inference} (also referred to as {\em graph structure learning}) is a problem arising in many applications of graph neural networks and geometric deep learning \cite{bronstein2021geometric}. Existing LGI approaches belong to two broad classes. {\bf Known input graph:} these approaches assume the input graph provided but imperfect in some sense, and attempt to modify it to improve the message passing. Various graph rewiring approaches fall in this category \cite{topping2022understanding,sun2022selforganization, chen2020iterative,jin2020prognn}.   {\bf Unknown input graph:} methods in this class learn the graph structure in a dynamic fashion, without necessarily assuming a fixed graph at the beginning of training. Some  approaches assume the latent graph to be {\em fully connected} (complete), e.g., Transformers  \cite{vaswani2017attention} and attentional multi-agent predictive models \cite{hoshen2017vain}, whereas other approaches perform {\em sparse} latent graph inference, offering computational and learning advantages. Notable sparse LGI techniques are LDS-GNN \cite{franceschi2019discretestruc}, Dynamic GCNNs \cite{wang2020dyncnn}, and variants of DGM \cite{kazi2022dgm,borde2023latent}. 

\paragraph{Topological deep learning}
or TDL stems from the pioneering works on Topological Signal Processing (TSP)  \cite{barbarossa2020topological, schaub2021signal, roddenberry2022cellsp, sardellitti2022cell} that showed the benefits of considering higher-order (multi-way) relationships among data points. Generalizations of the renowned Weisfeiler-Lehman graph isomorphism test to simplicial (SC) \cite{bodnar2021weisfeiler} and cell (CW) \cite{bodnarcwnet} complexes  have been proposed, along with SC and CW message-passing architectures. Convolutional SC and CW architectures have been previously studied in \cite{ebli2020simplicial, yang2022effsimpl, hajij2020cell, yang2023convolutional, roddenberry2021principled, hajij2022}. Attentional SC and CW architectures have been presented in \cite{giusti22, anonymous2022SAT, giusti2022cell}. A notable unifying framework for TDL was proposed in \cite{hajij2023topological}, where the concept of combinatorial complex (CC)   generalizing SCs, CWs, and hypergraphs, was introduced along with a general class of message-passing CC neural networks. An excellent survey on TDL can be found in \cite{papillon2023architectures}. Finally, Sheaf Neural Networks (SNNs) working on cellular sheaves \cite{hansen2019toward} built upon graphs were proposed in \cite{hansen2020sheaf, sheaf2022, battiloro2022tangent, battiloro2023tangent, barbero2022sheaf} and shown to be effective in heterophilic settings. 

Our paper is related to both classes of works: we improve and generalize DGM to perform latent topology inference and learning of non-regular topologies via the machinery of Topological Deep Learning \cite{hajij2023topological} (TDL).

\section{Background}

\paragraph{Regular cell complexes}
We start with the fundamentals of {\em regular cell complexes}, topological spaces that provide an effective way to represent complex interaction systems of various orders. Regular cell complexes generalize both graphs and simplicial complexes. 

\label{cell_section}
\textit{\textbf{Definition 1 (Regular cell complex) }}~\cite{hansen2019toward, bodnarcwnet}. 
A \textit{regular cell complex} (CW) is a topological space $\mathcal{C}$ together with a partition $\{\mathcal{X}_{\sigma}\}_{\sigma \in \mathcal{P}_{\mathcal{C}}}$ of subspaces $\mathcal{X}_{\sigma}$ of $\mathcal{C}$ called $\mathbf{cells}$, where $\mathcal{P}_{\mathcal{C}}$ is the indexing set of $\mathcal{C}$, s.t.

\begin{enumerate}[leftmargin=*]
    \item For each $c$ $\in$  $\mathcal{C}$, every sufficient small neighborhood of $c$ intersects finitely many $\mathcal{X}_{\sigma}$;  
    \item For all $\mathcal{\tau}$,$\mathcal{\sigma}$ we have that $\mathcal{X}_{\tau}$ $\cap$ $\overline{\mathcal{X}}_{\sigma}$ $\neq$ $\varnothing$ iff $\mathcal{X}_{\tau}$ $\subseteq$ $\overline{\mathcal{X}}_{\sigma}$, where $\overline{\mathcal{X}}_{\sigma}$ is the closure of the cell;
    \item Every $\mathcal{X}_{\sigma}$ is homeomorphic to $\mathbb{R}^{k}$ for some $k$;
    \item For every $\sigma$ $\in$ $\mathcal{P}_{\mathcal{C}}$ there is a homeomorphism $\phi$ of a closed ball in $\mathbb{R}^{k}$ to $\overline{\mathcal{X}}_{\sigma}$ such that the restriction of $\phi$ to the interior of the ball is a homeomorphism onto $\mathcal{X}_{\sigma}$.
\end{enumerate}

Condition 2 implies that the indexing set $\mathcal{P}_{\mathcal{C}}$ has a poset structure, given by $\tau$ $\leq$ $\sigma$ iff $\mathcal{X}_{\tau}$ $\subseteq$ $\overline{\mathcal{X}_\sigma}$, and we say that $\tau$ \textit{bounds} $\sigma$. This is known as the \textit{face poset} of $\mathcal{C}$. The regularity condition 4 implies that all of the topological information about $\mathcal{C}$ is encoded in the poset structure of $\mathcal{P}_{\mathcal{C}}$. Then, a regular cell complex can be identified with its face poset. For this reason, from now on we will indicate the cell $\mathcal{X}_{\sigma}$ with its corresponding face poset element $\sigma$. The dimension or order $\textrm{dim}(\sigma)$ of a cell $\sigma$ is $k$, we call it a $k-$cell and denote it with $\sigma^k$ to make this explicit when necessary. Regular cell complexes can be described via an incidence relation (boundary relation) with a  reflexive and transitive closure that is consistent with the partial order introduced in Definition 1.

\noindent\textit{\textbf{Definition 2 (Boundary Relation).}} We have the boundary relation  $\sigma$ $\prec$ $\tau$ iff $\dim({\sigma})$ $\leq$ $\dim({\tau})$ and there is no cell $\delta$ such that $\sigma$ $\leq$ $\delta$ $\leq$ $\tau$.

In other words, Definition 2 states that the boundary of a cell $\sigma^k$ of dimension $k$ is the set of all cells of dimension less than $k$ bounding $\sigma^k$. The dimension or order of a cell complex is the largest dimension of any of its cells. A graph $\mathcal{G}$ is a particular case of a cell complex of order 1, containing only cells of order 0 (nodes) and 1 (edges). We can use the previous definitions to introduce the four types of (local) adjacencies present in regular cell complexes: 

\textit{\textbf{Definition 3 (Cell Complex Adjacencies) }~\cite{bodnarcwnet}. For a cell complex $\mathcal{C}$ and a cell $\sigma \in \mathcal{P}_{\mathcal{C}}$, we define: }

\begin{itemize}[leftmargin=*]
    \item The boundary adjacent cells $\mathcal{B}(\sigma)$ $=$ $\{ \tau $ $|$ $\tau \prec \sigma \}$, are the lower-dimensional cells on the
boundary of $\sigma$. For instance, the boundary cells of an edge are its endpoint nodes.

    \item The co-boundary adjacent cell $\mathcal{CB}(\sigma)$ $=$ $\{\tau $  $|$ $  \sigma \prec \tau \}$, are the higher-dimensional cells with
    $\sigma$ on their boundary. E.g., the co-boundary cells of a node are the edges having that node as an endpoint.
    
    \item The lower adjacent cells $\mathcal{N}_{d}(\sigma)$ $=$ $\{ \tau $ $|$ $ \exists \delta$ such that $\delta \prec \sigma$ and $\delta \prec \tau\}$, are the cells of
the same dimension as $\sigma$ that share a lower dimensional cell on their boundary. The line graph
adjacencies between the edges are a typical example of this.

     \item The upper adjacent cells $\mathcal{N}_{u}(\sigma)$ $=$ $\{ \tau $ $|$ $ \exists \delta$ such that $\sigma \prec \delta$ and $\tau \prec \delta\}$. These are the cells of the same dimension as $\sigma$ that are on the boundary of the same higher-dimensional cell. 
\end{itemize}

\textit{\textbf{Definition 4 (k-skeleton)}}. A {\it k-skeleton} of a regular cell complex $\mathcal{C}$ is the subcomplex of $\mathcal{C}$ consisting of cells of dimension at most $k$.

From Definition 1 and Definition 5, it is clear that the 0-skeleton of a cell complex is a set of vertices and the 1-skeleton is a set of vertices and edges, thus a graph. For this reason, given a graph $\mathcal{G}=\{\mathcal{V},\mathcal{E}\}$, it is possible to build a regular cell complex "on top of it", i.e. a cell complex $\mathcal{C}_\mathcal{G}$ whose 1-skeleton is isomorphic to $\mathcal{G}$. 

\textbf{Remark 1.} We remark that the term ``regular'' is ambiguous in this context, because a regular cell complex is an object as in Definition 1, while a regular graph is a graph whose neighborhoods cardinalities are all the same. For this reason, in the following we will refer to regular cell complex(es) simply as ``cell complex(es),'' reserving the term ``regular'' only for graphs.

In this work, we attach order 2 cells to an inferred (learned) subset of induced cycles of the graph $\mathcal{G}$ up to length $K_{max}$, we refer to them as \textit{polygons}, and we denote the resulting order-2 cell complex with $\mathcal{C}_{\mathcal{G}}=\{\mathcal{V},\mathcal{E},\mathcal{P}\}$, where $\mathcal{P}$ is the polygons set, with $|\mathcal{V}| = N$, $|\mathcal{E}| = E$, and $|\mathcal{P}| = P$. This procedure is formally a skeleton-preserving lifting map; a detailed discussion about the lifiting of a graph into a cell complex can be found in \cite{bodnarcwnet}.

\paragraph{Signals over Cell Complexes} 
A $k$-cell signal is defined as a mapping from the set $\mathcal{D}_k$ of all $k$-cells contained in the complex, with $|\mathcal{D}_k|=N_k$, to real numbers \cite{sardellitti2022cell}:
\begin{equation}\label{signals}
     x_{k}: {\cal D}_{k} \rightarrow \mathbb{R}
\end{equation}
Therefore, for a order-2 complex $\mathcal{C}_{\mathcal{G}}$, the $k$-cell signals are defined as the following mappings: 
\begin{equation}
x_{0}: {\cal V} \rightarrow \mathbb{R} , \qquad x_{1}: {\cal E} \rightarrow \mathbb{R} , \qquad x_2: {\cal P} \rightarrow \mathbb{R} ,
\end{equation}
representing node, edge and polygon signals, respectively, with $N_0 = N$, $N_1=E$, and $N_2 = P$. If $F$ $k-$cell signals are available, we collect them in a feature matrix $\mathbf{X}_{k,in}=\{\mathbf{x}_{k,in}(i)\}_{i=1}^{N_k} \in \mathbb{R}^{N_k \times F}$, where $\mathbf{x}_{k,in}(i) = [x_1(\sigma^k_i), \dots, x_F(\sigma^k_i)]\in \mathbb{R}^{F}$ are the features of the $i$-th 
 $k-$cell $\sigma^k_i$. 

\textbf{Remark 2.} The definition of $k-$cell signal in \eqref{signals} is sufficient and rigorous for the scope of this paper, however more formal topological characterizations in terms of chain and cochain spaces can be given \cite{sardellitti2022cell,bodnarcwnet, hajij2023topological}.

\section{A Differentiable Layer for Latent Topology Inference}

\begin{figure}[t]
  \centering
  \includegraphics[width=1\columnwidth]{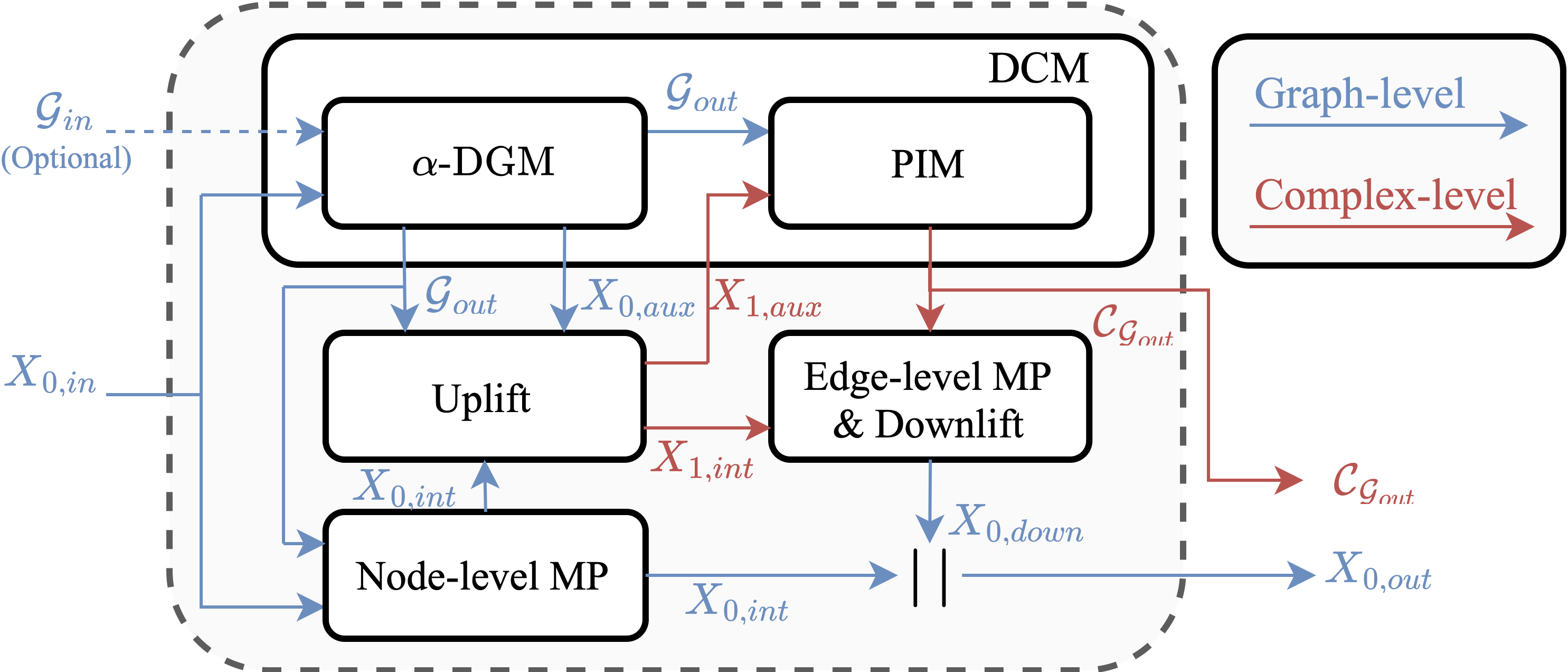}
  \caption{The DCM and the proposed architecture}
  \label{fig:dcm_det}
\end{figure}

We propose a novel layer, depicted at high level in Figure \ref{fig:dcm_easy} and in detail in Figure \ref{fig:dcm_det}, comprising of a series of modules among which the most important one is the \textit{Differentiable Cell Complex Module} (DCM), the first fully differentiable module that performs LTI, i.e. learns a cell complex describing the hidden  higher-order interactions among input data points. The DCM is equipped with a novel sampling algorithm that, at the best of our knowledge, is the first graph/complex sampling technique that allows to generate topologies whose neighborhoods cardinality is not fixed a priori but can be freely learned in an end-to-end fashion. 

The proposed layer takes as input the nodes feature matrix $\mathbf{X}_{0,in} \in \mathbb{R}^{N \times F_{in}}$, and gives as output the updated nodes feature matrix $\mathbf{X}_{0,out} \in \mathbb{R}^{N \times F_{out}}$ and the inferred latent cell complex $\mathcal{C}_{\mathcal{G}_{out}}$. Optionally, the layer can take as input also a graph $\mathcal{G}_{in}$. To make the layer self-consistent in a multi-layer setting, the output can be reduced to the  updated nodes feature matrix $\mathbf{X}_{0,out}$ and the 1-skeleton (graph) $\mathcal{G}_{out}$ of $\mathcal{C}_{\mathcal{G}_{out}}$.  

We employ a two-step inference procedure to keep the computational complexity  tractable, i.e. the DCM module first samples the 1-skeleton of the latent cell complex (possibly in a sparse way, i.e. $E << N^2$), and then it samples the polygons among the induced cycles generated by the sampled edges. The first step is implemented via the novel $\alpha$-Differentiable Graph Module ($\alpha-$DGM), while the second step is implemented via the novel Polygon Inference Module (PIM). Directly sampling among all the possible polygons, thus trivially generalizing the DGM framework \cite{kazi2022dgm}, would have lead to intractable complexity, e.g. even at triangles level the sampling would have had a complexity of the order of $\mathcal{O}(\sqrt{E^3})=\mathcal{O}(N^3)$, being all edges candidate to be sampled. In the following, we describe in details each module of the proposed layer.

\subsection{The $\alpha$-Differentiable Graph Module} 
 The $\alpha$-DGM is a novel variation of the DGM and it is responsible of inferring the 1-skeleton (graph) of the latent cell complex. One of the main limitations of the DGM \cite{kazi2022dgm} is the constraint of inferring only regular graphs; we solve this problem by proposing a novel sampling procedure based on the $\alpha-$entmax class of functions \cite{peters2019entmax}, recently introduced in the NLP community to obtain sparse
alignments and zero output probabilities. With fixed $\alpha>1$ and input $\mathbf{s} \in \mathbb{R}^d$, the $\alpha-$entmax function $\boldsymbol{\alpha}_{ENT}:\mathbb{R}^d \rightarrow \Delta^d$ is defined as: 
\begin{equation}
    \boldsymbol{\alpha}_{ENT}(\mathbf{s}) = \argmax_{\mathbf{u} \in \Delta^d} \;\mathbf{u}^\top\mathbf{s} + H_\alpha(\mathbf{u}),
    \label{eq:entmax}
\end{equation}
where $\Delta^d$ is the $d$-dimensional simplex and $H_\alpha$ is the Tsallis $\alpha$-entropy \cite{peters2019entmax}. For $\alpha=1$, equation \eqref{eq:entmax} recovers the standard softmax, while for $\alpha>1$ solutions can be sparse with a degree depending on $\alpha$. In practice, $\alpha$ can be initialized at a reasonable value (e.g., 1.5) and adapted via gradient descent \cite{correia-etal-2019-adaptively}.

In $\alpha$-DGM, we first compute auxiliary node features $\mathbf{X}_{0, aux} = \nu_{0}(\mathbf{X}_{0,in}) \in \mathbb{R}^{N \times d_0}$, where $\nu_{0}(\cdot)$ is a learnable function, e.g. a GNN if an initial graph $\mathcal{G}_{in}$ is provided or an MLP otherwise.  

At this point, a similarity function $\textrm{sim}(\cdot)$ is chosen, the similarities among node embeddings are computed and collected in the vectors $\mathbf{z}_i = \{\textrm{sim}(\mathbf{x}_{0,aux}(i),\mathbf{x}_{0,aux}(j))\}_j \in \mathbb{R}^N$, $i,j = 1, \dots, N$. 

The (sparse) edge probabilities are then obtained node-wise via $\alpha-$entmax, i.e. the vectors $\mathbf{p}_i = \boldsymbol{\alpha}_{ENT}(\mathcal{LN}(\mathbf{z}_i)) \in \mathbb{R}^N$, $i = 1,\dots,N$ are computed; layer normalization $\mathcal{LN}$ is employed to have better control on the similarities statistics and, consequently, more stability in the training procedure. The $(i,j)$ edge is included in the inferred graph if $\mathbf{p}_i(j)>0$. This procedure leads to a directed graph that would be incompatible with the cell complex structure introduced in Section \ref{cell_section}, whose 1-skeleton is an undirected graph; the inferred directed graph is converted to the closest undirected graph $\mathcal{G}_{out} = \{\mathcal{N},\mathcal{E}\}$, i.e.  each inferred edge is considered as a bidirectional edge to obtain $\mathcal{E}$. We want further stress the fact that computing edge probabilities via the $\alpha$-entmax leads to sparse and non-regular graphs whose sparsity level can be controlled (or learned) tuning the parameter $\alpha$. A detailed pictorial description of the $\alpha$-DGM is shown in Figure 3 (left).

\noindent \textbf{Remark 3.} The choice of applying the $\alpha$-entmax in a node-wise fashion and not directly on a vector containing all the possible similarities (that would have naturally lead to an undirected graph) is due to the fact that, in the latter case, we empirically observed that the $\alpha$-entmax consistently leads to heavily disconnected graphs with few hub nodes (therefore, also leading to performance drops). 

At this point,  the intermediate node features $\mathbf{X}_{0,int} \in \mathbb{R}^{N \times F_{int}}$ are computed with one or more usual message passing (MP) rounds over the inferred graph:
\begin{align} \label{int_node}
        \mathbf{x}_{0,int}(i) = \gamma_0\Bigg(\mathbf{x}_{0,in}(i), \bigoplus_{j \in \mathcal{N}_u(\sigma_i^0)}\phi_0^{\mathcal{N}_u}\Big(\mathbf{x}_{0,in}(i),\mathbf{x}_{0,in}(j)\Big)\Bigg),
    \end{align}
where $\gamma_0(\cdot)$ and $\phi_0^{\mathcal{N}_u}(\cdot)$ are lernable functions and $\bigoplus$  is any (possibly) permutation invariant operator.

\begin{figure}[t]
  \centering
  \includegraphics[width=.845\columnwidth]{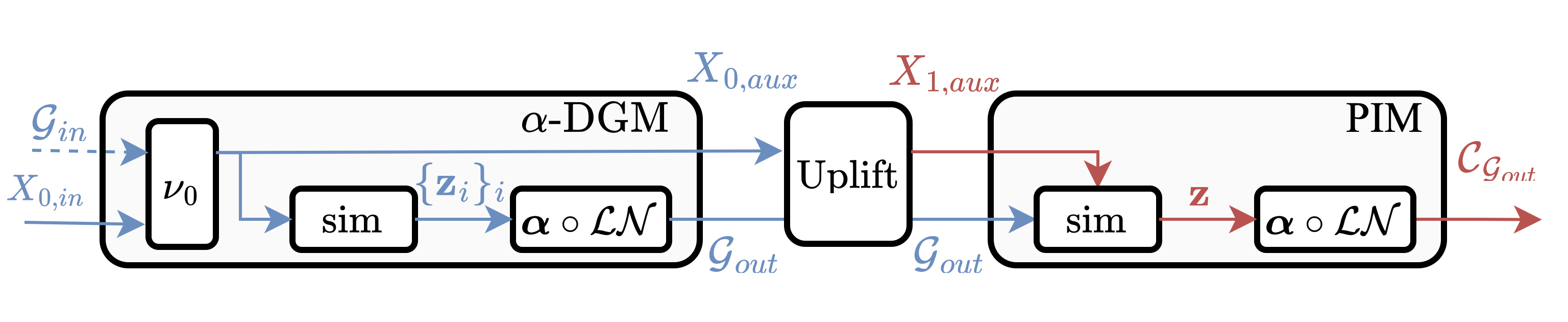}
  \caption{The $\alpha$-Differentiable Graph Module (left) and the Polygon Inference Module (right).}
\end{figure}

\subsection{Uplift Module}
To fully exploit the potential of cell complexes, it is not sufficient to work at the node level. For this reason, intermediate edge features are learned (or computed)  with a MP round of the form:
\begin{align} \label{int_edge}
        &\mathbf{x}_{1,int}(i) = \beta_{1}\Bigg(\bigoplus_{j \in \mathcal{B}(\sigma_i^1)}\phi^{\mathcal{B}}_1\Big(\mathbf{x}_{0,int}(j)\Big)\Bigg),
\end{align}
where $\beta_{1}(\cdot)$ is a (possibly) learnable function. Therefore, the intermediate  features $\mathbf{x}_{1,int}(i)$ of edge $i$ having as endpoints nodes $j$ and $v$ are learned (or computed) as a function of the intermediate node features $\mathbf{x}_{0,int}(j)$ and $\mathbf{x}_{0,int}(v)$.

\subsection{The Polygon Inference Module}
The Polygon Inference Module (PIM) is a novel module responsible of inferring the polygons of the latent cell complex by sampling a subset of the induced cycles of the inferred 1-skeleton  $\mathcal{G}_{out}$. 

To this aim, auxiliary edge features  are computed from the auxiliary node features with a MP round as in \eqref{int_edge}, which we denote in this context with $\mathbf{X}_{1, aux} = \nu_{1}(\mathbf{X}_{0,aux}) \in \mathbb{R}^{E \times d_1}$ for notation consistency. 

At this point, we need a similarity function for the edge embeddings belonging to the same induced cycle. We decide to use the sum of the pairwise similarities, e.g. for a triangle made by the edges $i$, $j$, $v$, the similarity is computed (with a slight abuse of notation) as:
\begin{align}\label{sim_poly}
    \textrm{sim}(\mathbf{x}_{1,aux}(i), \mathbf{x}_{1,aux}(j), \mathbf{x}_{1,aux}(v)) &= \textrm{sim}(\mathbf{x}_{1,aux}(i), \mathbf{x}_{1,aux}(j))+\textrm{sim}(\mathbf{x}_{1,aux}(i), \mathbf{x}_{1,aux}(v))\nonumber \\
    &+\textrm{sim}( \mathbf{x}_{1,aux}(j), \mathbf{x}_{1,aux}(v))
\end{align}
In general, the similarity in \eqref{sim_poly} for an induced cycle of length $k\leq K_{max}$ contains $\frac{k!}{(k-2)!2!}$ terms, however $K_{max}$ is usually a very small integer and the computation can be trivially distributed. 

The similarities are  collected in a vector $\mathbf{z}\in \mathbb{R}^{\widetilde{P}}$, where $\widetilde{P}$ is the number of induced cycles, and the polygons probabilities are computed as $\mathbf{p} = \boldsymbol{\alpha}_{ENT}(\mathcal{LN}(\mathbf{z})) \in \mathbb{R}^{\widetilde{P}}$. We set $\mathcal{C}_{\mathcal{G}_{out}} = \{\mathcal{N}, \mathcal{E}, \mathcal{P}\}$, where $\mathcal{P}$ are the induced cycles with positive probabilities and $|\mathcal{P}| = P \leq \widetilde{P}$ (possibly $P <<\widetilde{P}$). 

The updated edge features $\mathbf{X}_{1,out} \in \mathbb{R}^{E \times F_{out}}$ are computed with one or more MP rounds as:
\begin{align} \label{out_edge}
        \mathbf{x}_{1,out}(i) = \gamma_1\Bigg(\mathbf{x}_{1,int}(i),& \bigoplus_{j \in \mathcal{N}_u(\sigma^1_i)}\phi_1^{\mathcal{N}_u}\Big(\mathbf{x}_{1,int}(i),\mathbf{x}_{1,int}(j)\Big), \nonumber \\
        &\bigoplus_{j \in \mathcal{N}_d(\sigma^1_i)}\phi_1^{\mathcal{N}_d}\Big(\mathbf{x}_{1,int}(i),\mathbf{x}_{1,int}(j)\Big)\Bigg),
    \end{align}
where $\gamma_1(\cdot)$, $\phi_1^{\mathcal{N}_d}(\cdot)$, and $\phi_1^{\mathcal{N}_u}(\cdot)$ are learnable functions. A detailed pictorial description of the PIM is shown in Figure 3 (right).

\textbf{Remark 4.} The message passing rounds in \eqref{out_edge} and  in \eqref{int_edge} (as a special case) are instances of message passing neural networks over cell complexes. Several MP schemes can be defined for cell (simplicial, combinatorial) complexes, please refer to \cite{bodnarcwnet, hajij2023topological} for more details. We employ the schemes in \eqref{int_edge}-\eqref{out_edge} for two reasons: the first one is that using more sophisticated MP rounds (e.g. moving to cells of higher order than polygons or designing more intensive messages exchange among different cell orders) would have lead to computational intractability; the second one is that moving to higher order cells also introduces a series of tricky theoretical issues (in terms of the structure of the complex) that are not trivial to tackle \cite{hansen2019toward}. We plan to investigate these directions in future works.

\subsection{Downlift Module and Output Computation} 
At this point, the output node features $\mathbf{X}_{0,out} \in \mathbb{R}^{N \times F_{out}}$ are learned (or computed) from the updated edge features $\mathbf{X}_{1,out} \in \mathbb{R}^{E \times F_{out}}$ and the intermediate node features $\mathbf{X}_{0,int} \in \mathbb{R}^{N \times F_{int}}$ as:
\begin{align} \label{out_node}
        \mathbf{x}_{0,out}(i) = \Big[\mathbf{x}_{0,int}(i) \,\Big|\Big|\, 
        \mathbf{x}_{0,down}(i)\Big],
\end{align}
where the $\mathbf{X}_{0,down}\in \mathbb{R}^{N \times F_{out}}$ are obtained with a MP round of the form:

\begin{equation}
   \mathbf{x}_{0,down}(i) =  \beta_0\Bigg(\bigoplus_{j \in \mathcal{CB}(\sigma_i^{0})}\phi_{0}^{\mathcal{CB}}\Big(\mathbf{x}_{1,out}(j)\Big)\Bigg)
\end{equation}
with $\beta_{0}(\cdot)$ being a (possibly) learnable function. Therefore, the output features $\mathbf{x}_{0,out}(i)$ of node $i$ are learned (or computed) as the concatenation of its intermediate features and features obtained downlifiting the updated features of the edges for which $i$ is an endpoint. 

We present a comparison in terms of computational complexity between the DCM  and the DGM \cite{kazi2022dgm} in Appendix B. 

\subsection{Training of the Differentiable Cell Module}
The proposed sampling scheme based on the $\alpha$-entmax does not allow the gradient of the downstream task loss
to flow both through the graph and polygons inference  branches of the DCM, due to the fact that they involves only auxiliary features and the entmax outputs are substantially just a way of indexing the edges and the polygons present in the inferred complex. To enable a task-oriented end-to-end training of the DCM, we generalize the approach of DGM \cite{kazi2022dgm,borde2023latent} and design an additional loss term
that rewards edges and polygons involved in a correct classification (in classification tasks) or in "good" predictions (in regression tasks), and
penalizes edges and polygons that lead to misclassification or high error predictions. Refer to Appendix A for the details.

\section{Experimental Results} \label{sec:experiments}
In this Section, we evaluate the effectiveness of the proposed framework on several heterophilic and homophilic  graph benchmarks. Briefly, homophily is the principle that similar users interact at a higher rate than dissimilar ones. In a graph, this means that nodes with similar characteristics are more likely to be connected. Traditional Graph/Cell Complex Convolutional Neural Networks (GCNs and CCCNs) implicitly rely on the homophily assumption, and one typically observes performance drops of these models in heterophilic settings \cite{sheaf2022,spinelli2021fairdrop}.

Our main goal is to show that {\em latent topology inference} via the differentiable cell complex module (DCM) allows learning higher-order relationships among data points that lead to significant improvements w.r.t. {\em latent graph inference}.  Since DCM is a generalization of the differentiable graph module (DGM), we use as a comparison the original (discrete) DGM \cite{kazi2022dgm} (denoted DGM-E) and its recently introduced non-Euclidean version   \cite{borde2023latent} (denoted DGM-M). Moreover, we also report the results of a simplified variant of our model (denoted as DCM ($\alpha=1$)), in which the graph is explicitly learned and all the polygons are taken, i.e. $\alpha = 1$ in the PIM; in this (lower complexity) case, the $\alpha$-DGM guides both the edges (explicitly) and the polygons (implicitly) inference steps. Finally, we report the vanilla MLP and GCN as further baselines. In all the experiments, we utilize GCNs at the graph level and CCCNs at the edge level; the details about the model architectures employed to obtain the presented results are given in Appendix E. The \textcolor{red}{\textbf{first}} and the \textcolor{blue}{\textbf{second}} best results  are highlighted.

\begin{table}[t]
  \caption{Homophilic graph node classification benchmarks. Test accuracy in \% 
  avg.ed over 10 splits.}
  \label{tab:res-homo}
  \centering
\scalebox{.815}{\begin{tabular}{ll ccccc}
    \hline & & Cora & CiteSeer & PubMed & Physics & CS \\
    \hline
  &  Model/Hom. level\hspace{-6mm} & 0.81 & 0.74 & 0.80 & 0.93 & 0.80 \\
    
    \hline
    \parbox[t]{2.5mm}{\multirow{5}{*}{\rotatebox[origin=c]{90}{w/o graph}}}
   &  MLP & 58.92 $\pm$ 3.28 & 59.48 $\pm$ 2.14 & 85.75 $\pm$ 1.02 & 94.91 $\pm$ 0.28 & 87.80 $\pm$ 1.54 \\
   & DGM-E \cite{kazi2022dgm} & 62.48 $\pm$ 3.24 & 62.47 $\pm$ 3.20 & 83.89 $\pm$ 0.70 & 94.03 $\pm$ 0.45 & 76.05 $\pm$ 6.89 \\
   & DGM-M \cite{borde2023latent} & \textcolor{black}{\text{70.85}} $\pm$ 4.30 & \textcolor{black}{\text{68.86}} $\pm$ 2.97 & \textcolor{blue}{\textbf{87.43}} $\pm$ 0.40 & \textcolor{black}{\text{95.25}} $\pm$ 0.36 & \textcolor{black}{\text{92.22}} $\pm$ 1.09 \\
   \cline{2-7}
   &  DCM & \textcolor{red}{\textbf{78.80}} $\pm$ 1.84 & \textcolor{red}{\textbf{76.47}} $\pm$ 2.45 & \textcolor{black}{\text{87.38}} $\pm$ 0.91 & \textcolor{red}{\textbf{96.45}} $\pm$ 0.12 & \textcolor{red}{\textbf{95.40}} $\pm$ 0.40 \\
    & DCM ($\alpha=1$) & \textcolor{blue}{\textbf{78.73}} $\pm$ 1.99 & \textcolor{blue}{\textbf{76.32}} $\pm$ 2.75 & \textcolor{red}{\textbf{87.47}} $\pm$ 0.77 & \textcolor{blue}{\textbf{96.22}} $\pm$ 0.27 & \textcolor{blue}{\textbf{95.35}} $\pm$ 0.37 \\ 
%
    \hline
        \parbox[t]{2.5mm}{\multirow{5}{*}{\rotatebox[origin=c]{90}{w graph}}} &
    GCN & 83.11 $\pm$ 2.29 & 69.97$\pm$ 2.00 & 85.75 $\pm$ 1.01 & 95.51 $\pm$ 0.34 & 87.28$\pm$ 1.54\\
   & DGM-E \cite{kazi2022dgm} & 82.11 $\pm$ 4.24 & 72.35 $\pm$ 1.92 & 87.69 $\pm$ 0.67 & 95.96 $\pm$ 0.40 & 87.17 $\pm$ 3.82 \\
   & DGM-M \cite{borde2023latent} & \textcolor{red}{\textbf{86.63}} $\pm$ 3.25 & \textcolor{black}{\text{75.42}} $\pm$ 2.39 & \textcolor{black}{\text{87.82}} $\pm$ 0.59 & \textcolor{black}{\text{96.21}} $\pm$ 0.44 & \textcolor{black}{\text{92.86}}  $\pm$ 0.96 \\
   \cline{2-7}
   &  DCM & \textcolor{black}{\text{85.78}} $\pm$ 1.71 & \textcolor{red}{\textbf{78.72}} $\pm$ 2.84 & \textcolor{blue}{\textbf{88.49}} $\pm$ 0.62 & \textcolor{red}{\textbf{96.99}} $\pm$ 0.44 & \textcolor{red}{\textbf{95.79}} $\pm$ 0.48 \\
   &  DCM ($\alpha=1$) & \textcolor{blue}{\textbf{85.97}} $\pm$ 1.86 & \textcolor{blue}{\textbf{78.60}} $\pm$ 3.16 & \textcolor{red}{\textbf{88.61}} $\pm$ 0.69 & \textcolor{blue}{\textbf{96.69}} $\pm$ 0.46 & \textcolor{blue}{\textbf{95.78}} $\pm$ 0.49 \\

    \bottomrule
    
    \end{tabular}}
    
\end{table}

\begin{table}[t]\vspace{-.5cm}
  \caption{Heterophilic graph node classification benchmarks. Test accuracy in \% 
  avg.ed over 10 splits.}
  \label{tab:res-hetero}
  \centering
\scalebox{.815}{\begin{tabular}{ll cccc}
\toprule & & Texas & Wisconsin & Squirrel & Chameleon \\
\midrule
& Model/Hom. level\hspace{-6mm} & 0.11 & 0.21 & 0.22 & 0.23 \\

\midrule
        \parbox[t]{2.5mm}{\multirow{5}{*}{\rotatebox[origin=c]{90}{w/o graph}}} &
MLP & 77.78 $\pm$ 10.24 & 85.33 $\pm$ 4.99 & 30.44 $\pm$ 2.55 & 40.35 $\pm$ 3.37 \\
& DGM-E \cite{kazi2022dgm} & 80.00 $\pm$ 8.31  & \textcolor{blue}{\textbf{88.00}} $\pm$ 5.65  & 34.35 $\pm$ 2.34  & 48.90 $\pm$ 3.61 \\
& DGM-M \cite{borde2023latent} & 81.67 $\pm$ 7.05 & \textcolor{red}{\textbf{89.33}} $\pm$ 1.89 & 35.00 $\pm$ 2.35 & 48.90 $\pm$ 3.61 \\ 
\cline{2-6}
& DCM & \textcolor{red}{\textbf{85.71}} $\pm$ 7.87 & 87.49 $\pm$ 5.94 & \textcolor{red}{\textbf{35.55}} $\pm$ 2.24 & \textcolor{blue}{\textbf{53.63}} $\pm$ 3.07\\
& DCM ($\alpha=1$) & \textcolor{blue}{\textbf{84.96}} $\pm$ 10.24 & 86.72 $\pm$ 6.02 & \textcolor{blue}{\textbf{35.25}} $\pm$ 2.22 & \textcolor{red}{\textbf{53.67}} $\pm$ 3.19 \\

\midrule
    \parbox[t]{2.5mm}{\multirow{5}{*}{\rotatebox[origin=c]{90}{w graph}}} &
GCN & 41.66 $\pm$ 11.72 & 47.20 $\pm$ 9.76 & 24.19 $\pm$ 2.56 & 32.56 $\pm$ 3.53\\
& DGM-E \cite{kazi2022dgm} & 60.56 $\pm$ 8.03 & 70.67 $\pm$ 10.49 & 29.87 $\pm$ 2.46 & 44.19 $\pm$ 3.85 \\
& DGM-M \cite{borde2023latent} & 62.78 $\pm$ 9.31 & 76.00 $\pm$ 3.26 & 30.44 $\pm$ 2.38 & 45.68 $\pm$ 2.66 \\ 
\cline{2-6}
& DCM & \textcolor{blue}{\textbf{84.87}} $\pm$ 10.04 & \textcolor{red}{\textbf{86.33}} $\pm$ 5.14 & \textcolor{blue}{\textbf{34.95}} $\pm$ 2.59  & \textcolor{blue}{\textbf{53.05}} $\pm$ 3.00 \\
& DCM ($\alpha=1$) & \textcolor{red}{\textbf{84.96}} $\pm$ 5.60 & \textcolor{blue}{\textbf{85.36}} $\pm$ 5.05 & \textcolor{red}{\textbf{35.13}} $\pm$ 2.27  & \textcolor{red}{\textbf{53.76} $\pm$ 3.72} \\
\bottomrule
\end{tabular}}

\end{table}

We follow the same core experimental set up of \cite{borde2023latent} on transductive classification tasks; in particular, we first focus on standard graph datasets such as {\em Cora}, {\em CiteSeer} \cite{yang2016revisitingsemisup}, {\em PubMed}, {\em Physics} and {\em CS} \cite{shchur2019pitfalls}, which have high homophily levels ranging from 0.74 to 0.93. We then test our method on several challenging heterophilic datasets, {\em Texas}, {\em Wisconsin}, {\em Squirrel}, and {\em Chameleon}  \cite{rozemberczki2021hetero}, which have low homophily levels ranging between 0.11 and 0.23. The results for the homophilic and heterophilic datasets are presented in Tables~\ref{tab:res-homo} and~\ref{tab:res-hetero}, respectively. All the models were  tested in two settings: assuming the original graph is available (marked {\em w graph} in Tables~\ref{tab:res-homo} and~\ref{tab:res-hetero}) and a more challenging case in which the input graph is assumed to be unavailable ({\em w/o graph}). 

From Tables~\ref{tab:res-homo} and~\ref{tab:res-hetero}, it is evident that the DCM exhibits consistently superior performance compared to alternative methods across both homophilic and heterophilic datasets, both with and without provided input graphs. As expected, our method achieves top performance when the graph is not provided for the heterophilic datasets and when the graph is provided for the homophilic datasets. Somewhat surprising results are achieved when considering the input graph for heterophilic datasets. Despite the model being given ``wrong'' input graphs (in the sense that the structure of the graph does not match the structure of the node features), the overall performance of the DCM remains stable, yielding remarkable improvements exceeding $20\%$ for Texas and approximately $10\%$ for Wisconsin compared to DGM-M \cite{borde2023latent}. Therefore, observing the results across both homophilic and heterophilic datasets when the input graph is provided, we can appreciate the fact that the DCM is capable of exploiting the available ``good'' graph in homophilic cases to improve performance, while also being less sensitive to the available ``wrong'' graph in the case of heterophilic datasets.

\begin{figure*}[t]
  \centering
  \begin{subfigure}[b]{0.319\linewidth}
    \centering
    \includegraphics[width=\textwidth]{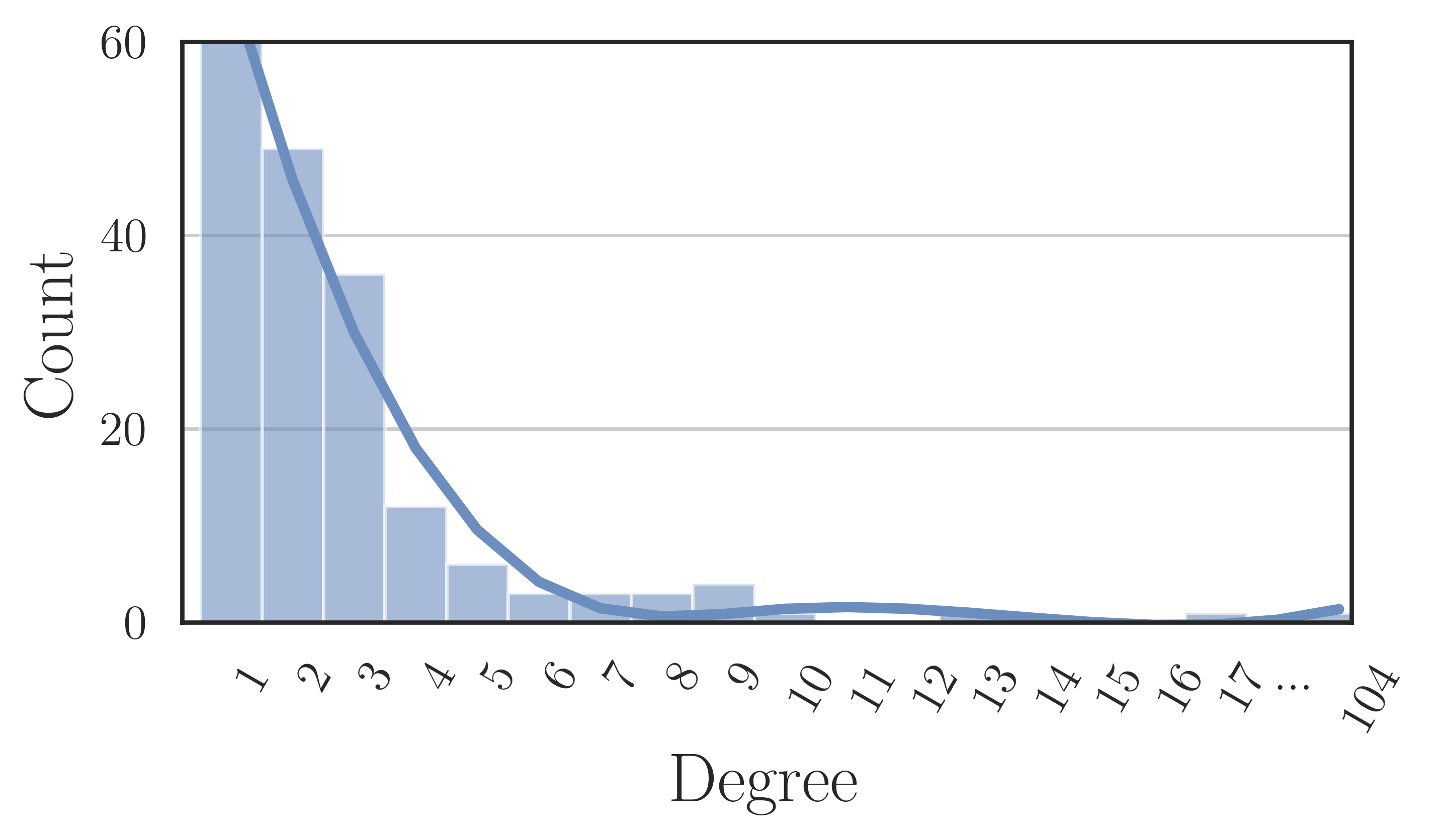}
    \vspace{-8pt}
  \end{subfigure}
  \begin{subfigure}[b]{0.32\linewidth}
    \centering
    \includegraphics[width=\textwidth]{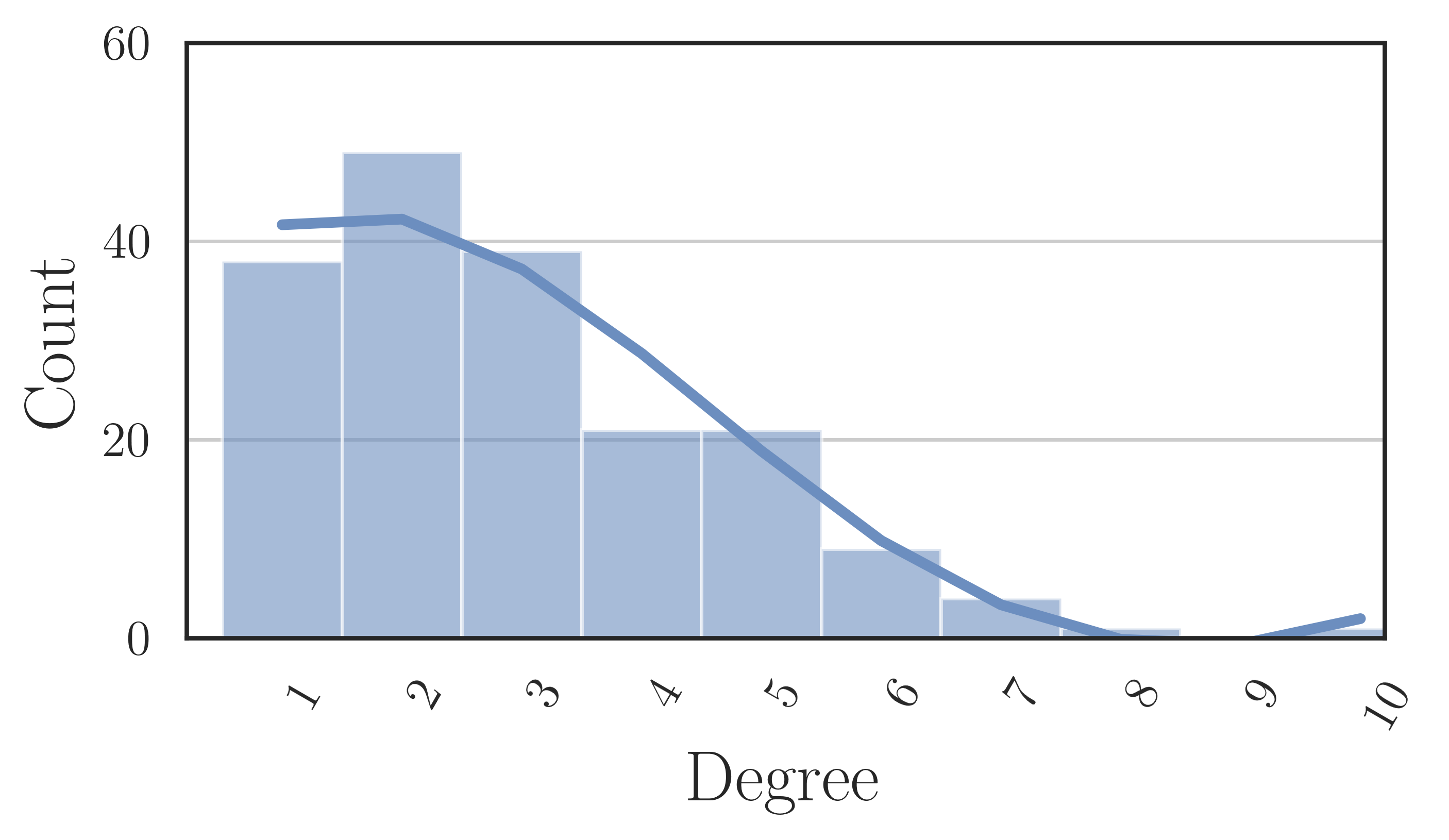}
    \vspace{-8pt}
  \end{subfigure}
  \begin{subfigure}[b]{0.322\linewidth}
    \centering
    \includegraphics[width=\textwidth]{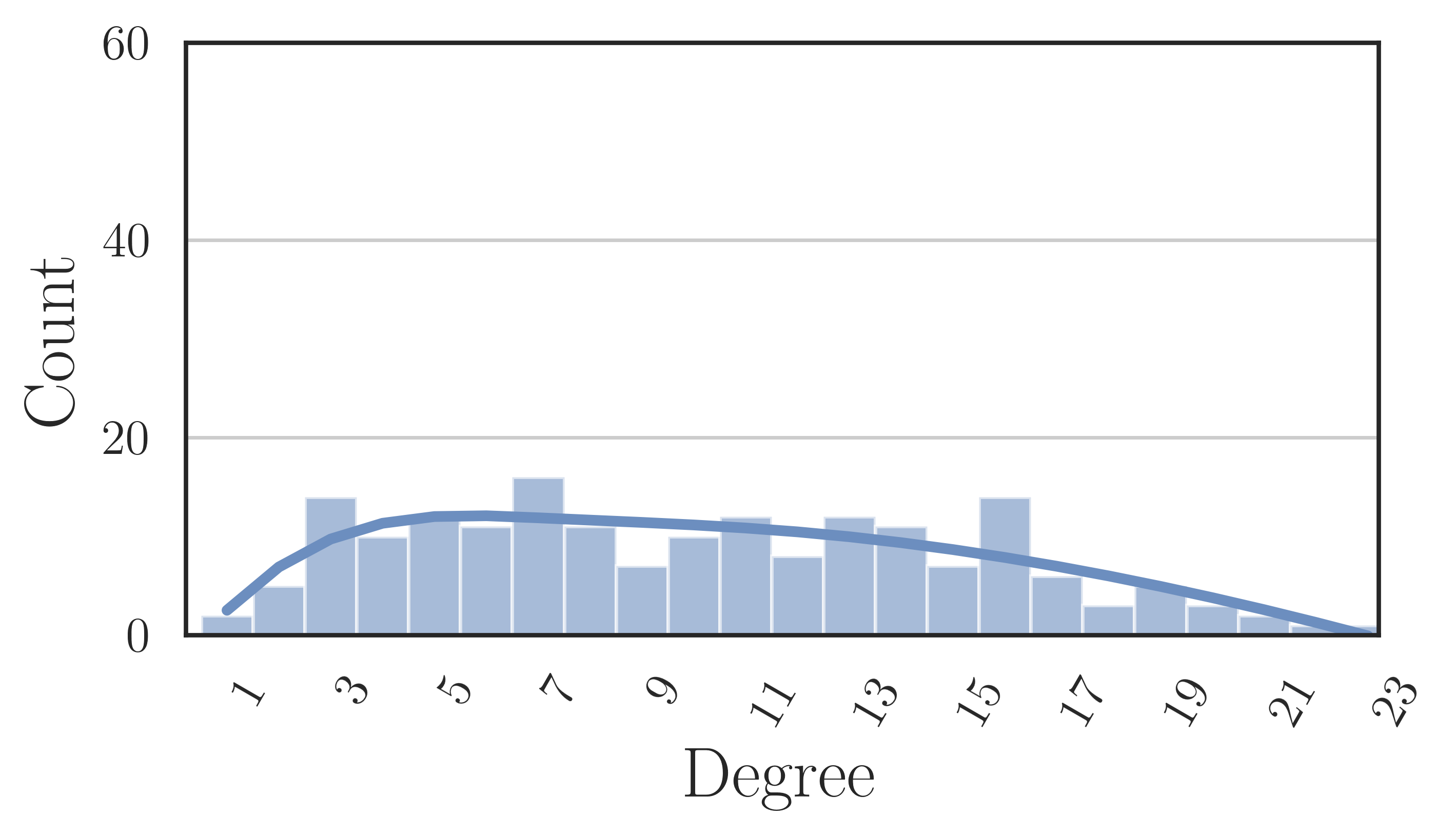}
    \vspace{-8pt}
  \end{subfigure}
  
  \vskip 0.1cm

  \centering
  \begin{subfigure}[b]{0.32\linewidth}
    \centering
    \includegraphics[width=\textwidth]{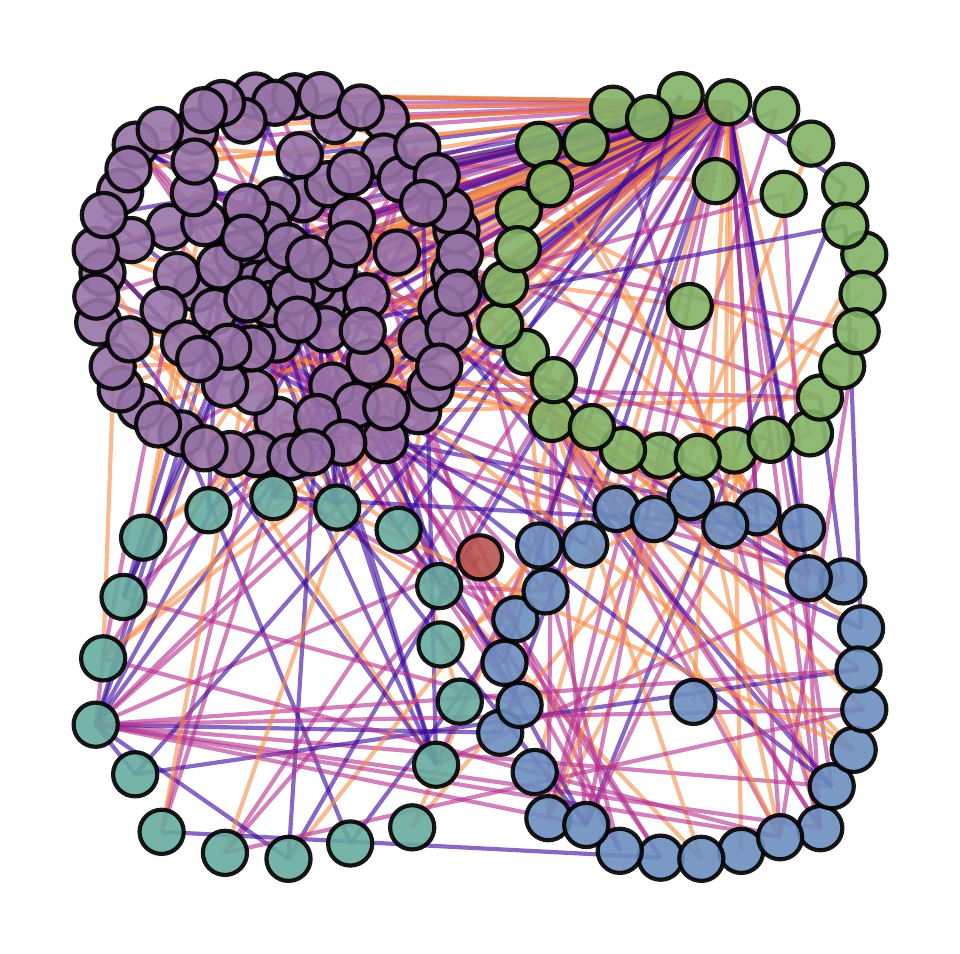}
    \vspace{-8pt}
    \caption{Input, $\%_{p} = 100$, $h=0.11$}
  \end{subfigure}
  \begin{subfigure}[b]{0.32\linewidth}
    \centering
    \includegraphics[width=\textwidth]{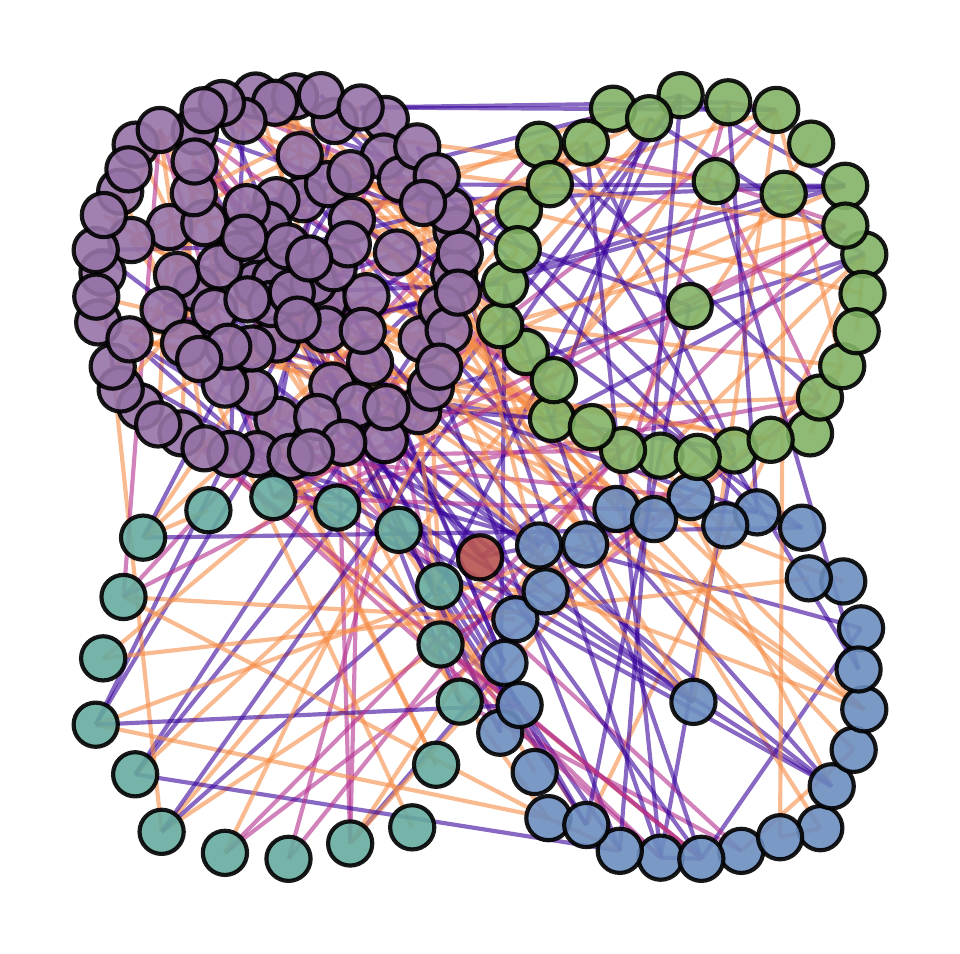}
    \vspace{-8pt}
    \caption{Epoch 40, $\%_{p} = 90 $, $h=0.4$}
  \end{subfigure}
  \begin{subfigure}[b]{0.32\linewidth}
    \centering
    \includegraphics[width=\textwidth]{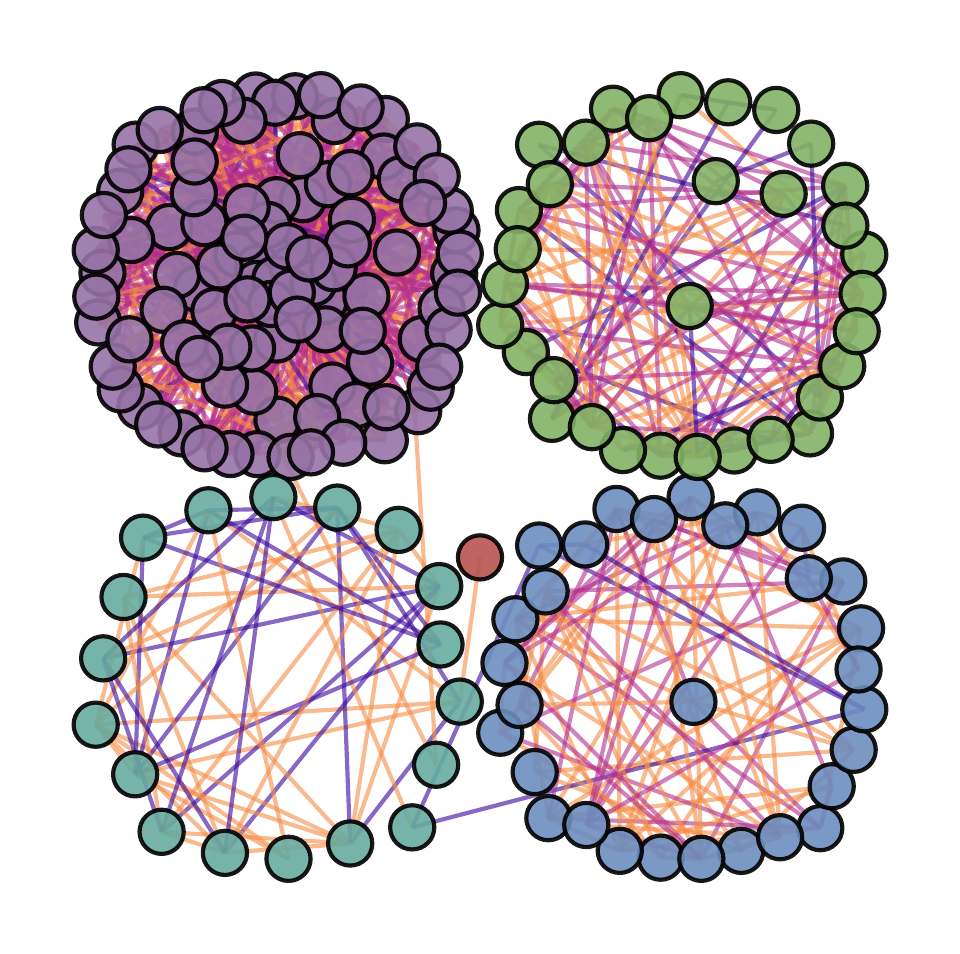}
    \vspace{-8pt}
    \caption{Epoch 180, $\%_{p} = 20 $, $h=0.99$}
  \end{subfigure}

    \caption{Evolution of the latent complex for the Texas dataset, along with homophily level and nodes degree distribution. Edges in \textcolor{edge_col}{orange}, triangles in \textcolor{triangle_col}{lilac}, squares in \textcolor{square_col}{purple} ($K_{max} = 4$)}
    \label{fig:plots}
\end{figure*}

The good performance on the heterophilic datasets suggests that the learned latent complex has a homophilic  1-skeleton (graph), that enables the involved GCNs and CCCNs to reach higher accuracies. To corroborate this hypotheses, in Figure \ref{fig:plots} we show the evolution of the latent topology during the training for the Texas dataset along with the nodes degree distribution, the percentage of sampled polygons $\%_{p}$, and the homophily level $h$ that we note reaching $0.99$ on the final inferred graph from the initial $0.11$. Moreover, the fact that most of the inferred polygons belong to the same class and the pretty high spread of the degree distribution further confirm the effectiveness of the proposed architecture. More experiments, ablation studies, and plots can be found in Appendix C and D.

\section{Conclusions} 
We introduced the paradigm of latent topology inference, aiming to not (only) learn a latent graph structure but rather higher-order topological structures describing multi-way interactions among data-points. We made LTI implementable by introducing the Differentiable Cell Complex Module, a novel learnable function that dynamically learns a regular cell complex to improve the downstream task in an end-to-end fashion and in a computationally tractable way, via to a two-step inference procedure that avoids an exhaustive search across all possible cells in the input. We showed the effectiveness of the DCM on a series of homophilic and heterophilic graph benchmarks dataset, comparing its performance against state-of-the-art latent graph inference methods, and showing its competitive performance both when the input graph is provided or not.

\section{Limitations and Future Directions}
To our knowledge, DCM is the first differentiable approach for latent topology inference. The  promising results open several avenues for future work. 
%
\paragraph{Methodological.} Although cell complexes are very  flexible topological objects, other instances of LTI could leverage hypergraphs or combinatorial complexes (CCs) \cite{hajij2022}, which are able to handle both hierarchical and set-type higher-order interactions. Second, remaining within cell complexes, different MP schemes \cite{hajij2023topological}, lifting maps \cite{bodnarcwnet}, or model spaces \cite{borde2023latent} are of interest for future work. 
Third, 
extending our framework to weighted \cite{battiloro2023weighted} or directed complexes \cite{owen2018directedcomplex} could give further insights. 
Finally, one potentially interesting direction is merging LTI and Sheaf Neural Networks in order to learn cellular sheaves defined on higher-order complexes and not just fixed and given  graphs. 

\paragraph{Computational.} 
While most of our experimental validation of the DCM focused on transductive tasks, we could also tackle inductive ones \cite{borde2023latent}. Second,  though computationally tractable thanks to the proposed two-step inference procedure, DCM may benefit from further improvements to effectively scale on very large datasets. The main bottlenecks are MP operations and the search for the induced cycles of the learned graph. A possible solution for the former could be  neighbor samplers \cite{borde2023latent}, while the latter could be tackled by leveraging  stochastic search methods or moving to more flexible topological spaces, e.g. CCs. Finally, our sampling strategy implicitly assumes that there are no specific edges and polygons distributions requirements, e.g. a sampling budget is given or a particular correlation structure needs to be imposed on the cells. In these cases, incorporating more sophisticated sampling methods like IMLE \cite{li2018implicit,niepert2021imleexp,serra2022imlegraph} could be beneficial.

\section{Broader Impact}
The rise of GNNs applications in industrial domains, such as recommender systems, social networks, or computational chemistry has raised concerns about the potential adverse societal outcomes in case they are misused. Our research does not specifically examine these potential negative applications. Our main goal is to develop a framework that enrich the understanding of existing models and gives tools to discover and investigate interactions in data at various resolutions. We firmly believe that gaining a deeper understanding of machine learning models is essential for managing their societal implications and taking proactive measures to mitigate  potential negative effects.

\section{Disclosure of Funding}
This work was partly funded by the Sapienza grant RM1221816BD028D6 (DESMOS: DEsigning Self-explainable AI MOdels for Scientific applications), and by the PNRR MUR project PE0000013 (FAIR: Future Artificial Intelligence Research).

\bibliographystyle{IEEEbib}
\bibliography{biblio}
\medskip
\begin{appendix}
\newpage
\section{Training Procedure}

Here we discuss the key concepts about training of the DCM. We detail how to back-propagate through the entmax-based discrete sampling at the 1-cell (edge) level (in the $\alpha-$DGM) and at the 2-cell (polygon) level (in the PIM). We follow the approach proposed by\cite{kazi2022dgm} and introduce a supplementary loss that rewards edges and polygons involved in a correct classification and penalizes the ones which result in misclassification. We define the reward function:
\begin{equation}
    \delta\left(y_i, \hat{y}_i\right)=\mathbb{E}\left(a_i\right)-a_i \,,
\end{equation}
as the difference between the average accuracy of the i-th sample and the current prediction accuracy, where $y_i$ and $ \hat y_i$ are the predicted and true labels, and $a_i =1$ if $y_i = \hat y_i$ or 0 otherwise. We then define the loss associated to the edge sampling as follows:
\begin{equation}
    L_{G L}=\sum_{i=1}^N\left(\delta\left(y_i, \hat{y}_i\right) \sum_{j=1}^N \mathbf{p}_i(j)\right)
\end{equation}

We estimate $\mathbb{E}\left(a_i\right)$ with an exponential moving average on the accuracy  during the training process
\begin{equation}\label{loss_edge}
    \mathbb{E}\left(a_i\right)^{(t+1)}=\mu \mathbb{E}\left(a_i\right)^{(t)}+(1-\mu) a_i \,,
\end{equation}
with $\mu= 0.9$ in all our experiments and $\mathbb{E}\left(a_i\right)^{(0)} = \frac{1}{\text{\# classes}}$.

Similarly, the loss associated to the polygon sampling is defined as follows:
\begin{equation}\label{loss_poly}
    L_{P L}=\sum_{i=1}^N\left(\delta\left(y_i, \hat{y}_i\right) \sum_{j \in \mathcal{P}_i} \mathbf{p}(j)\right),
\end{equation}
where $\mathcal{P}_i$ is the set of indices of the polygons for which node $i$ is a vertex. Following \cite{borde2023latent}, the accuracy can be replaced with the R2 score in the case of regression tasks. In every experiment, we set the initial value of $\alpha$ to 1.5.

\section{Computational Complexity}
We carry out a complexity analysis of the proposed architecture with a particular focus on the differences with respect to the DGM \cite{kazi2022dgm}, which are:

(a) the introduction of $\alpha$-entmax to sample cells compared to the Gumbel-Softmax sampler; 

(b) the need to search for all cycles up to length $K_{max}$ in the skeleton $\mathcal{G}_{out}$;

(c) the sampling operation over the cycles;

(d) the cell complex neural network. 

For (a), the solution to \eqref{eq:entmax} cannot be expressed in closed form except for specific values of $\alpha$ \cite{correia-etal-2019-adaptively}, but an $\varepsilon$-approximate solution can be obtained in $\mathcal{O}(1/\log \varepsilon)$ time with a bisection algorithm \cite{blondel2019learning}. 

For (b), we leverage the algorithm in \cite{bodnarcwnet}, which has complexity $\mathcal{O}((E + N\widetilde{P}) \text{polylog}(N))$, where $N$ is the number of vertices, $E$ the number of sampled edges, and $\widetilde{P}$ the number of cycles induced by the skeleton up to length $K_{max}$. Note that in our implementation the cell complex is recomputed for each iteration, but this computation can be amortized across epochs, approximated with stochastic search algorithms, or leveraging more flexible topological spaces, e.g. Combinatorial Complexes.

For (c), the complexity is the same as (a), which is approximately linear in the number $\widetilde{P}$ of cycles. 

For (d), the complexity of message passing is also approximately linear in the size of the cell complex, due to the fact that  we consider cells of a constant maximum dimension and boundary size \cite{bodnarcwnet}.

\section{Additional Results}
\label{ap:add_res}
We present a series of additional experiments and ablation studies to further validate the effectiveness of the DCM. In particular, in Section \ref{ap:dgm} we investigate the advantages of employing the $\alpha$-entmax as sampling strategy w.r.t.  the Gumbel Top-k  trick used in the DGM \cite{borde2023latent,kazi2022dgm}. Additionally, in Section \ref{ap:stc}, we study the impact of the number of MP layers in \eqref{out_edge}-\eqref{int_node} (implemented, as explained in Section \ref{sec:experiments} of the body and Appendix E, with GCNs and CCCNs). Finally, in Section \ref{ap:max_cycle}, we investigate the usage of different values for the maximum cycle size $K_{max}$. 
\subsection{Sampling strategies}
\label{ap:dgm}
In this section, we assess the impact of employing the $\alpha$-entmax as sampling strategy, compared to the Top-k sampler utilized in \cite{kazi2022dgm,borde2023latent}, for both homophilic (Table \ref{tab:abl_sampl_homo}) and heterophilic datasets (Table \ref{tab:abl_sampl_hetero}), with and without input graph. We compare the two variants of the DCM (again denoted with DCM and DCM $(\alpha=1))$ against our same architecture but with Gumbel Top-k sampler in place of the $\alpha-$entmax, both when explicit sampling is performed at edge and polygons level (denoted with Top-k DCM) and when only the edge are sampled and all the polygons are taken (the counterpart of DCM $(\alpha=1)$, denoted with Top-K DCM All). Our investigation reveals that the capability of $\alpha$-entmax of generating non-regular topologies leads to significant performance gains on almost all the tested datasets.

\begin{table}[t]\vspace{-.5cm}
  \caption{Comparison among sampling strategies on homophilic graph node classification benchmarks. Test accuracy in \% 
  avg.ed over 10 splits.}
  \label{tab:abl_sampl_homo}
  \centering
\scalebox{.86}{\begin{tabular}{ll ccccc}
    \hline & & Cora & CiteSeer & PubMed & Physics & CS \\    
    \hline
    \parbox[t]{2.5mm}{\multirow{5}{*}{\rotatebox[origin=l]{90}{w/o graph}}} &
    DCM & \textcolor{black}{\textbf{78.80}} $\pm$ 1.84 & \textcolor{black}{\textbf{76.47}} $\pm$ 2.45 & \textcolor{black}{\text{87.38}} $\pm$ 0.91 & 96.45 $\pm$ 0.12 & \textcolor{black}{\textbf{95.40}} $\pm$ 0.40 \\
    & DCM ($\alpha=1$) & 78.73 $\pm$ 1.99 & 76.32 $\pm$ 2.75 & \textcolor{black}{\textbf{87.47}} $\pm$ 0.77 & 96.2 $\pm$ 0.27 & 95.35 $\pm$ 0.37 \\
   & Top-k DCM & 76.16 $\pm$ 3.71 & 73.21
 $\pm$ 2.73 & 87.13 $\pm$ 1.22 & \textbf{96.51} $\pm$ 0.49 & 95.25 $\pm$ 0.08 \\
   & Top-k DCM All & 76.24 $\pm$ 2.84 & 73.45
 $\pm$ 3.12 & 87.38 $\pm$ 1.04 & 96.49 $\pm$ 0.64 & 95.17 $\pm$ 0.07 \\
\midrule
    \parbox[t]{2.5mm}{\multirow{5}{*}{\rotatebox[origin=l]{90}{w graph}}} &
    DCM & \textcolor{black}{\textbf{85.78}} $\pm$ 1.71 & \textcolor{black}{\textbf{78.72}} $\pm$ 2.84 & 88.49 $\pm$ 0.62 & \textcolor{black}{\textbf{96.99}} $\pm$ 0.44 & \textcolor{black}{\textbf{95.79}} $\pm$ 0.48 \\
 
    & DCM ($\alpha=1$) & 85.97 $\pm$ 1.86 & 78.60 $\pm$ 3.16 & \textcolor{black}{\textbf{88.61}} $\pm$ 0.69 & 96.69 $\pm$ 0.46 & 94.81 $\pm$ 0.49 \\

  &  Top-k DCM & 76.96 $\pm$ 3.46 & 75.50
 $\pm$ 2.15 & 86.76 $\pm$ 0.89 & 96.19 $\pm$ 0.33 & 94.74 $\pm$ 0.28 \\
  &  Top-k DCM All & 77.55 $\pm$ 3.18 & 75.66
 $\pm$ 2.67 & 86.64 $\pm$ 0.74 & 96.21 $\pm$ 0.34 & 94.79 $\pm$ 0.37 \\
    \bottomrule
    
    \end{tabular}}
    
\end{table}

\begin{table}[t]\vspace{-.5cm}
  \caption{Comparison among sampling strategies on heterophilic graph node classification benchmarks. Test accuracy in \% 
  avg.ed over 10 splits.}
  \label{tab:abl_sampl_hetero}
  \centering
\scalebox{.86}{\begin{tabular}{ll cccc}
\toprule & & Texas & Wisconsin & Squirrel & Chameleon \\
\midrule
        \parbox[t]{1mm}{\multirow{5}{*}{\rotatebox[origin=l]{90}{w/o graph}}} &
DCM & \textcolor{black}{\textbf{85.71}} $\pm$ 7.87 & \textbf{87.49} $\pm$ 5.94 & \textcolor{black}{\textbf{35.55}} $\pm$ 2.24 & 53.63 $\pm$ 3.07\\
& DCM ($\alpha=1$) & 84.96 $\pm$ 10.24 & 86.72 $\pm$ 6.02 & 35.25 $\pm$ 2.22 & \textcolor{black}{\textbf{53.67}} $\pm$ 3.19 \\

&    Top-k DCM & 84.21 $\pm$ 8.21 & 84.10 $\pm$ 6.60 &  33.90 $\pm$ 1.38 & 52.88 $\pm$ 4.24 \\
&    Top-k DCM All &  82.71 $\pm$ 11.25  & 85.18 $\pm$ 6.89 & 33.80 $\pm$ 1.47 & 53.23 $\pm$ 3.86 \\
\midrule
    \parbox[t]{2.5mm}{\multirow{5}{*}{\rotatebox[origin=l]{90}{w graph}}} &
 DCM & 84.87 $\pm$ 10.04 & \textcolor{black}{\textbf{86.33}} $\pm$ 5.14 & 34.95 $\pm$ 2.59  & 53.05 $\pm$ 3.00 \\
& DCM ($\alpha=1$) & \textcolor{black}{\textbf{84.96}} $\pm$ 5.60 & 85.36 $\pm$ 5.05 & \textcolor{black}{\textbf{35.13}} $\pm$ 2.27  & \textcolor{black}{\textbf{53.76} $\pm$ 3.72} \\

 &   Top-k DCM & 84.61 $\pm$ 8.47 & 82.87	$\pm$ 7.59 & 33.77	$\pm$ 0.91 & 52.84	$\pm$ 4.24 \\
 &   Top-k DCM All & 84.66	$\pm$ 9.33 & 82.10	$\pm$ 6.64 & 33.52	$\pm$ 1.08 & 51.61	$\pm$ 4.07 \\
    \bottomrule
    
    \end{tabular}}
    
\end{table}

\subsection{Number of Message Passing Layers}
\label{ap:stc}
In this section, we assess the impact of the number of MP layers on the performance of our model. In Table \ref{tab:abl_MP_homo} (for homophilic datasets) and Table \ref{tab:abl_MP_hetero} (for heterophilic datasets), we show the accuracy as a function of the number of MP layers, i.e. GCNs and CCCNs layers at node and edge levels, respectively. We notice that the determination of an optimal number of message passing layers is not governed by a universal rule but rather depends on the characteristics of the dataset under consideration. However, we observe that employing a single message passing layer consistently yields favorable performance across most of the datasets, further confirming  that integrating higher-order information is beneficial. Moreover, the  1-layer configuration maintains a comparable number of trainable parameters w.r.t. the DGM-M \cite{borde2023latent} settings, whose results are reported as comparison. In particular, DGM-M  employs 3-layers GCNs   while we employ 1-layer GCNs and 1-layer CCCNs (having 3 times the number of parameters of a single GCN layer), see Appendix E for details. For this reason, despite the fact that some results in Table \ref{tab:abl_MP_homo} and in Table \ref{tab:abl_MP_hetero} are better than the one reported in the body of the paper, we decided to show the ones that correspond to architectures whose number of trainable parameters are as similar as possible to the reported competitors.

\begin{table}[t]\vspace{-.5cm}
  \caption{Results varying the number of MP layers on homophilic graph node classification benchmarks. Test accuracy in \% 
  avg.ed over 10 splits.}
  \label{tab:abl_MP_homo}
  \centering
\scalebox{.86}{\begin{tabular}{lc ccccc}
    \hline & \# MP layers & Cora & CiteSeer & PubMed & Physics & CS \\    
    \hline
    \parbox[t]{2.5mm}{\multirow{5}{*}{\rotatebox[origin=l]{90}{w/o graph}}} &
   1 & \textbf{78.80} $\pm$ 1.84 & \textbf{76.47} $\pm$ 2.45 & 87.38 $\pm$ 0.91 & 96.45 $\pm$ 0.12 & \textbf{95.40} $\pm$ 0.40 \\
    & 2 & 76.90 $\pm$ 2.81 & 75.86 $\pm$ 2.35 & 87.77 $\pm$ 0.53 & \textbf{97.10} $\pm$ 0.15  & 94.81 $\pm$ 0.40 \\
    & 3 & 74.98 $\pm$ 4.20 & 74.59 $\pm$ 2.35 & \textbf{88.10} $\pm$ 0.45 & 97.09 $\pm$ 0.11  & 94.15 $\pm$ 0.63 \\
    & 4 & 69.74 $\pm$ 3.95 & 71.41 $\pm$ 2.47 & 87.79 $\pm$ 0.36 & 96.80 $\pm$ 0.08  & 93.54 $\pm$ 0.72\\
    \midrule
    \parbox[t]{2.5mm}{\multirow{5}{*}{\rotatebox[origin=l]{90}{ w graph}}} &
   1 & 85.78 $\pm$ 1.71 & \textbf{78.72} $\pm$ 2.84 & 88.49 $\pm$ 0.62 & 96.99 $\pm$ 0.44 & \textbf{95.79} $\pm$ 0.48 \\
    & 2 & \textbf{88.89} $\pm$ 1.63 & 78.42 $\pm$ 2.63 & 88.27 $\pm$ 0.54 & \textbf{97.03} $\pm$ 0.23  & 94.50 $\pm$ 0.55 \\
    & 3 & 88.07 $\pm$ 2.26 & 76.95 $\pm$ 2.84 & \textbf{88.97} $\pm$ 0.61 & 97.01 $\pm$ 0.12  & 94.47 $\pm$ 0.54 \\
    & 4 & 86.00 $\pm$ 2.21 & 76.62 $\pm$ 2.86 & 88.62 $\pm$ 0.76 & 96.86 $\pm$ 0.19  & 93.66 $\pm$ 0.51\\

    \bottomrule
    
    \end{tabular}}
    
\end{table}
\begin{table}[t]\vspace{-.5cm}
  \caption{Results varying the number of MP layers on heterophilic graph node classification benchmarks. Test accuracy in \% 
  avg.ed over 10 splits.}
  \label{tab:abl_MP_hetero}
  \centering
\scalebox{.86}{\begin{tabular}{lc cccc}
\hline & \# MP layers & Texas & Wisconsin & Squirrel & Chameleon \\

\midrule
        \parbox[t]{1mm}{\multirow{5}{*}{\rotatebox[origin=l]{90}{w/o graph}}} &
1 & 85.71 $\pm$ 7.87 & \textbf{87.49} $\pm$ 5.94 & \textbf{35.55} $\pm$ 2.24 & \textbf{53.63} $\pm$ 3.07\\
& 2 & \textbf{87.89} $\pm$ 10.24 & 84.80 $\pm$ 4.81 & 33.30 $\pm$ 1.67 & 52.22 $\pm$ 4.22\\
& 3 & 82.02 $\pm$ 8.97 & 83.64 $\pm$ 4.87 & 33.61 $\pm$ 2.55 & 51.34 $\pm$ 3.25\\
& 4 & 80.75 $\pm$ 10.80 & 79.49 $\pm$ 7.45 & 31.96 $\pm$ 2.56 & 47.82 $\pm$ 3.11\\
\midrule
        \parbox[t]{1mm}{\multirow{5}{*}{\rotatebox[origin=l]{90}{w graph}}} &
1 & \textbf{84.87} $\pm$ 10.04 & \textbf{86.33} $\pm$ 5.14 & \textbf{34.95} $\pm$ 2.59 & \textbf{53.05} $\pm$ 3.00 \\
& 2 & 77.34 $\pm$ 8.64 & 76.31 $\pm$ 4.92 & 33.53 $\pm$ 2.24 & 52.61 $\pm$ 2.73\\
& 3 & 77.34 $\pm$ 8.09 & 80.95 $\pm$ 6.52 & 33.18 $\pm$ 1.88 & 51.47 $\pm$ 2.80\\
& 4 & 74.76 $\pm$ 12.01 & 76.13 $\pm$ 7.43 & 33.45 $\pm$ 2.59 & 43.65 $\pm$ 3.07\\
    \bottomrule
    
    \end{tabular}}
\end{table}

\begin{table}[h!]\vspace{-.5cm}
  \caption{Results varying $K_{max}$ on homophilic graph node classification benchmarks. Test accuracy in \% 
  avg.ed over 10 splits.}
  \label{tab:abl_kmax_homo}
  \centering
\scalebox{.86}{\begin{tabular}{lc ccccc}
    \toprule & $K_{max}$ & Cora & CiteSeer & PubMed & Physics & CS \\

    \hline

    \parbox[t]{2.5mm}{\multirow{3}{*}{\rotatebox[origin=l]{90}{\scriptsize{w/o graph }}}} &
    3 & 78.38 $\pm$ 2.50 & 75.56 $\pm$ 1.45 & 87.44 $\pm$ 0.91 & 96.32 $\pm$ 0.28 & \textbf{95.41} $\pm$ 0.60 \\
    & 4 & \textbf{78.80} $\pm$ 1.84 & \textbf{76.47} $\pm$ 2.45 & 87.38 $\pm$ 0.91 & 96.45 $\pm$ 0.12 & 95.40 $\pm$ 0.40 \\
    & 5 & 78.60 $\pm$ 2.60 & 75.62 $\pm$ 1.26 & \textbf{87.54} $\pm$ 0.98 & 96.20 $\pm$ 0.28 & 95.39 $\pm$ 0.60 \\
    \hline
    \parbox[t]{2.5mm}{\multirow{3}{*}{\rotatebox[origin=l]{90}{\scriptsize{w graph }}}} &
    3 & 85.54 $\pm$ 2.40 & 78.38 $\pm$ 1.56 & \textbf{88.50} $\pm$ 0.78 & 97.01 $\pm$ 0.28 & 95.68 $\pm$ 0.63 \\
    & 4 & \textbf{85.78} $\pm$ 1.71 & \textbf{78.72} $\pm$ 2.84 & 88.49 $\pm$ 0.62 & 96.99 $\pm$ 0.44 & \textbf{95.79} $\pm$ 0.48 \\
    & 5 & 85.54 $\pm$ 2.43 & 78.32 $\pm$ 1.47 & 85.75 $\pm$ 1.02 & \textbf{97.07} $\pm$ 0.40 & 95.68 $\pm$ 0.56 \\
    \bottomrule
    
    \end{tabular}}
    
\end{table}

\begin{table}[h!]\vspace{-.5cm}
  \caption{Results varying $K_{max}$ on heterophilic graph node classification benchmarks. Test accuracy in \% 
  avg.ed over 10 splits.}
  \label{tab:abl_kmax_hetero}
  \centering
\scalebox{.86}{\begin{tabular}{lc cccc}
    \toprule  & $K_{max}$ & Texas & Wisconsin & Squirrel & Chameleon \\
    \hline
    \parbox[t]{2.5mm}{\multirow{3}{*}{\rotatebox[origin=l]{90}{\scriptsize{w/o graph }}}} &
    3 &  85.53	$\pm$ 9.85  &  84.80 $\pm$	6.29 & 35.15 $\pm$	1.76 & 53.41 $\pm$	3.30 \\
    
    & 4 & \textbf{85.71} $\pm$ 7.87 & \textbf{87.49} $\pm$ 5.94 & \textbf{35.55} $\pm$ 2.24 & \textbf{53.63} $\pm$ 3.07\\

    & 5 &  80.22 $\pm$ 11.54  &  85.57 $\pm$	7.19 &  34.98 $\pm$	2.15 & 53.01 $\pm$	3.73 \\

    \hline
    \parbox[t]{2.5mm}{\multirow{3}{*}{\rotatebox[origin=l]{90}{\scriptsize{w graph }}}} &
    3 &  \textbf{85.49} $\pm$ 7.89  &  80.95 $\pm$	7.36 &  33.11 $\pm$	2.58 & 52.70 $\pm$	3.21 \\
    
    & 4 & 84.87 $\pm$ 10.04 & \textbf{86.33} $\pm$ 5.14 & \textbf{34.95} $\pm$ 2.59  & \textbf{53.05} $\pm$ 3.00 \\
    
    & 5 &  84.21 $\pm$ 9.15 &  81.71 $\pm$	6.33 &  33.88 $\pm$	1.49 & 52.92 $\pm$ 3.30 \\

    \bottomrule
    
    \end{tabular}}
    
\end{table}

\subsection{Maximum Cycle Size}\label{ap:max_cycle}
In this section, we conduct an ablation study on the maximum length $K_{max}$ of induced cycles taken in consideration to sample the polygons of the latent cell complex. In Table \ref{tab:abl_kmax_homo} and in Table \ref{tab:abl_kmax_hetero}, we show the accuracy as a function of $K_{max}$, respectively. As for the choice of the number of MP layers, even the optimal $K_{max}$ varies based on the specific dataset, however with top performance obtained in most cases when $K_{max}=4$.

\section{Plots}

\begin{figure*}[t]

    \hspace{1.5cm}\begin{subfigure}[b]{0.25\linewidth}
    \centering
    \includegraphics[width=\textwidth]{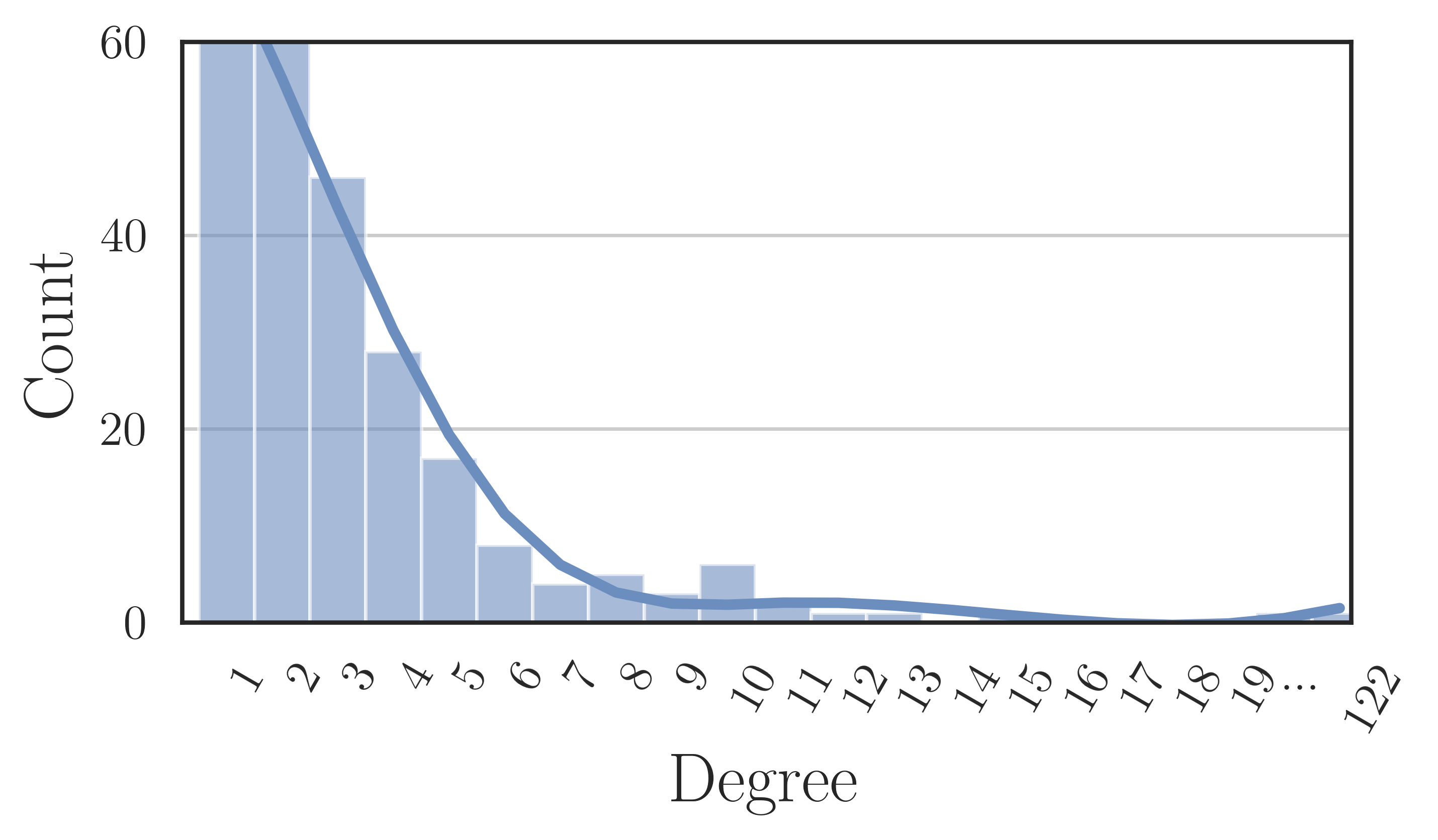}
    \vspace{-8pt}
  \end{subfigure}
  \begin{subfigure}[b]{0.25\linewidth}
    \centering
    \includegraphics[width=\textwidth]{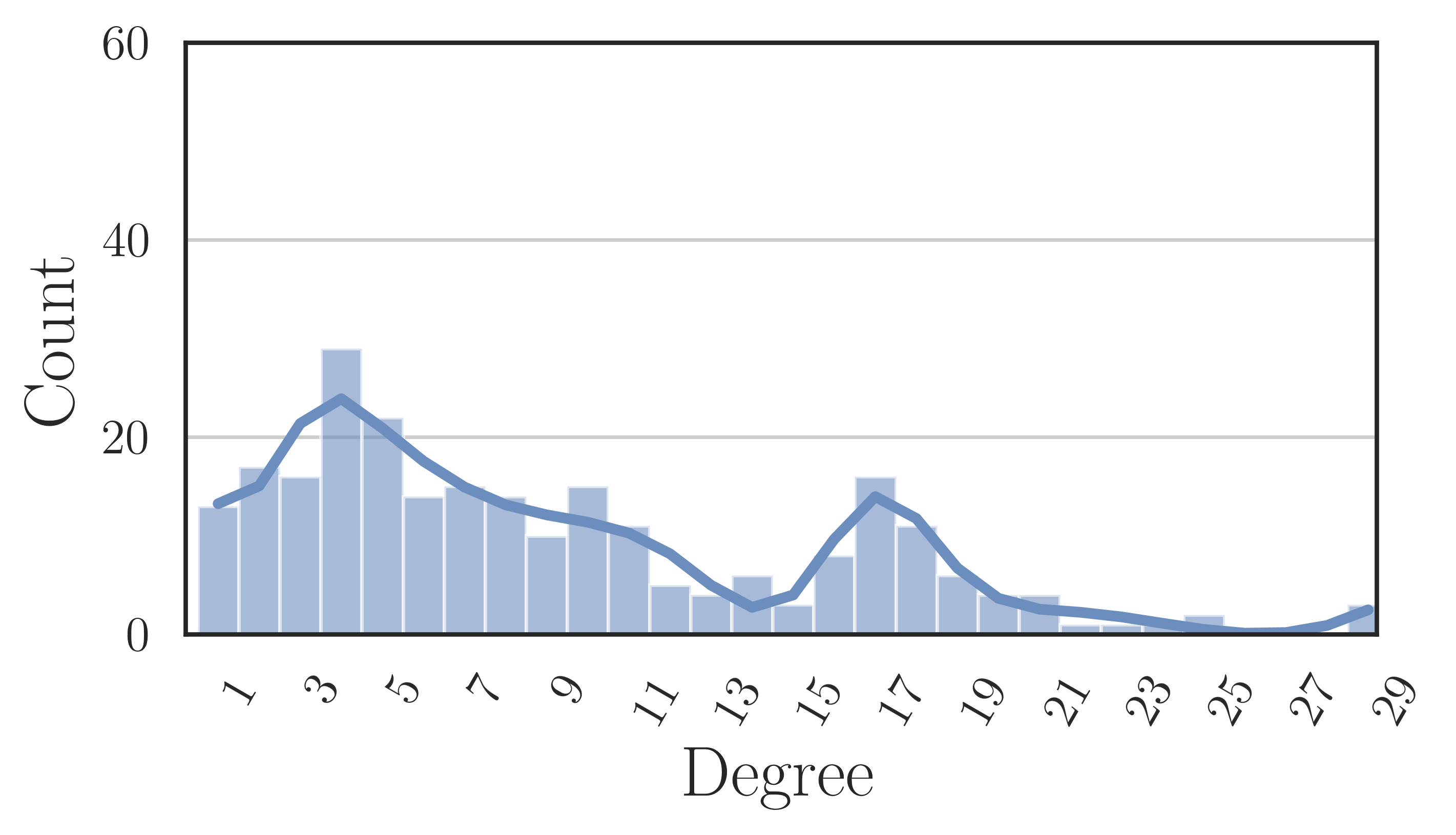}
    \vspace{-8pt}
  \end{subfigure}
  \begin{subfigure}[b]{0.25\linewidth}
    \centering
    \includegraphics[width=\textwidth]{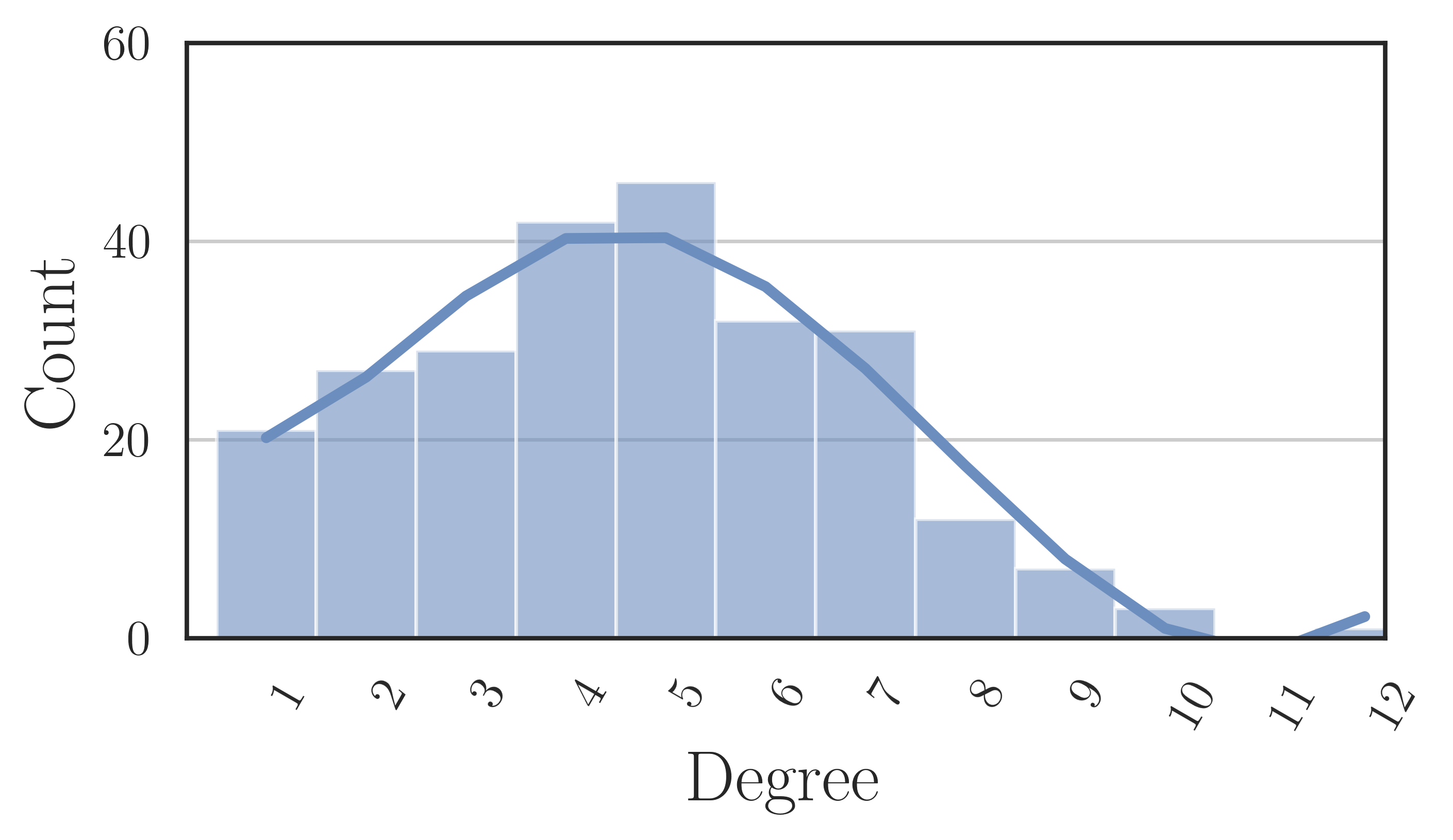}
    \vspace{-8pt}
    \end{subfigure}
    
  \centering
  \begin{subfigure}[b]{0.25\linewidth}
    \centering
    \includegraphics[width=\textwidth]{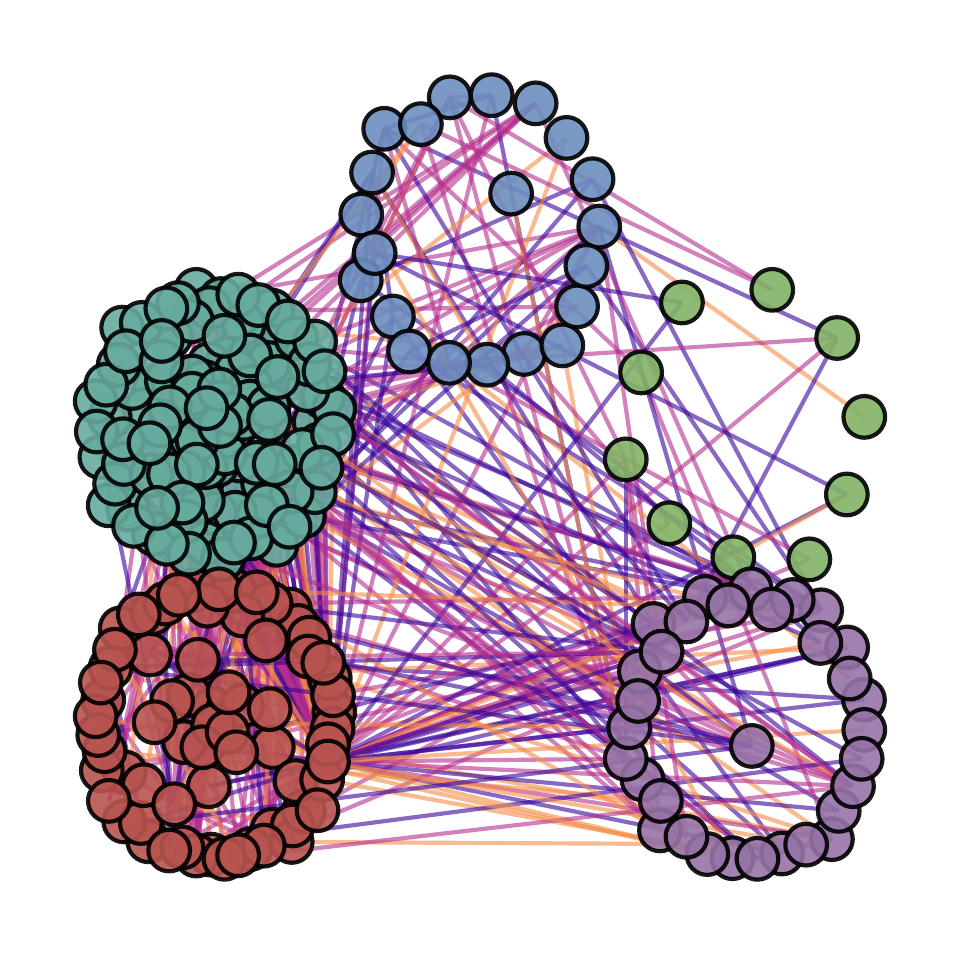}
    \vspace{-8pt}
    \caption{Input, $\%_{p} = 100$, $h=0.19$}
  \end{subfigure}
  \begin{subfigure}[b]{0.25\linewidth}
    \centering
    \includegraphics[width=\textwidth]{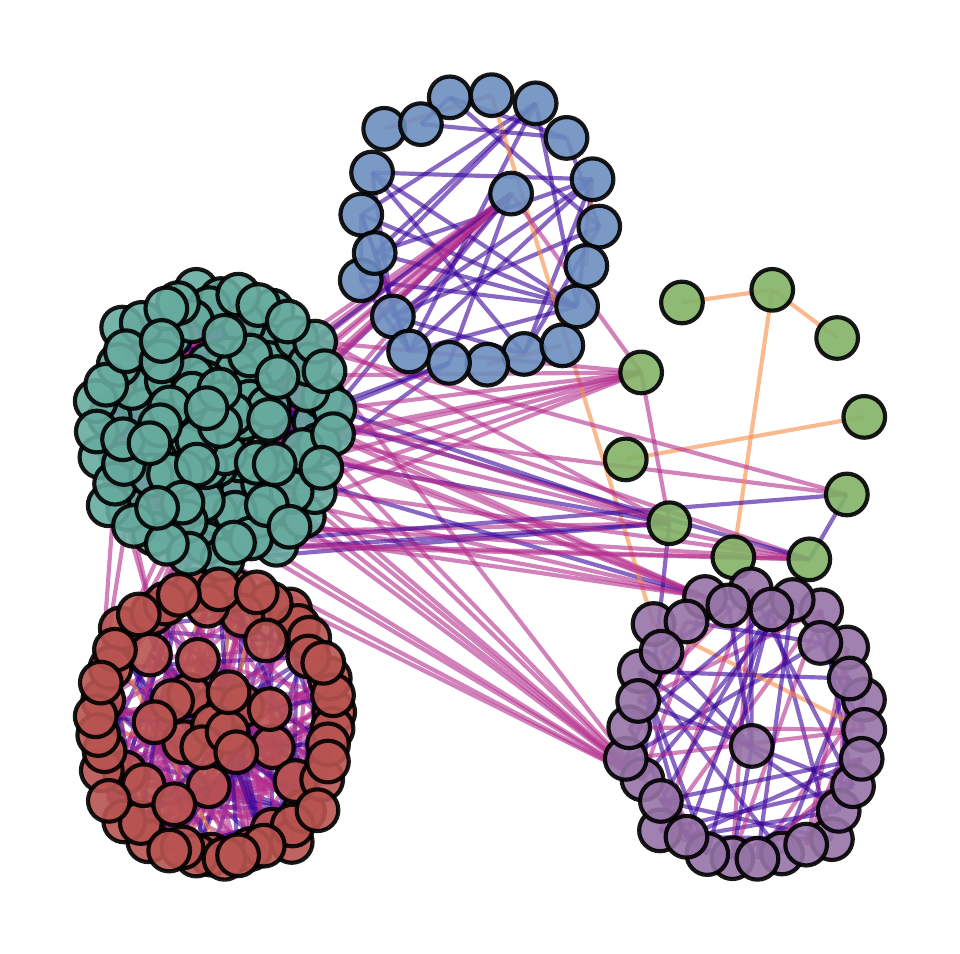}
    \vspace{-8pt}
    \caption{Epoch 10, $\%_{p} = 75 $, $h=0.70$}
  \end{subfigure}
  \begin{subfigure}[b]{0.25\linewidth}
    \centering
    \includegraphics[width=\textwidth]{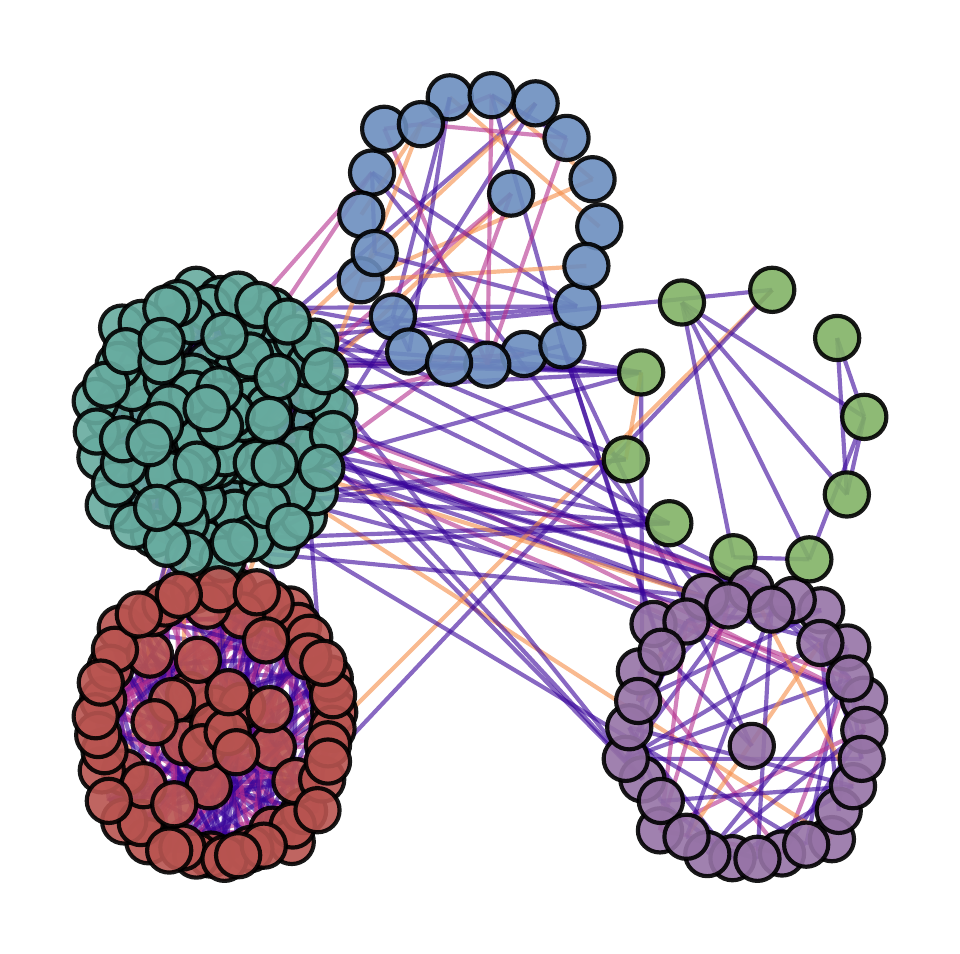}
    \vspace{-8pt}
    \caption{Epoch 50, $\%_{p} = 51 $, $h=0.89$}
  \end{subfigure}
  \vskip 1.0cm

  \begin{subfigure}[b]{0.25\linewidth}
    \centering
    \includegraphics[width=\textwidth]{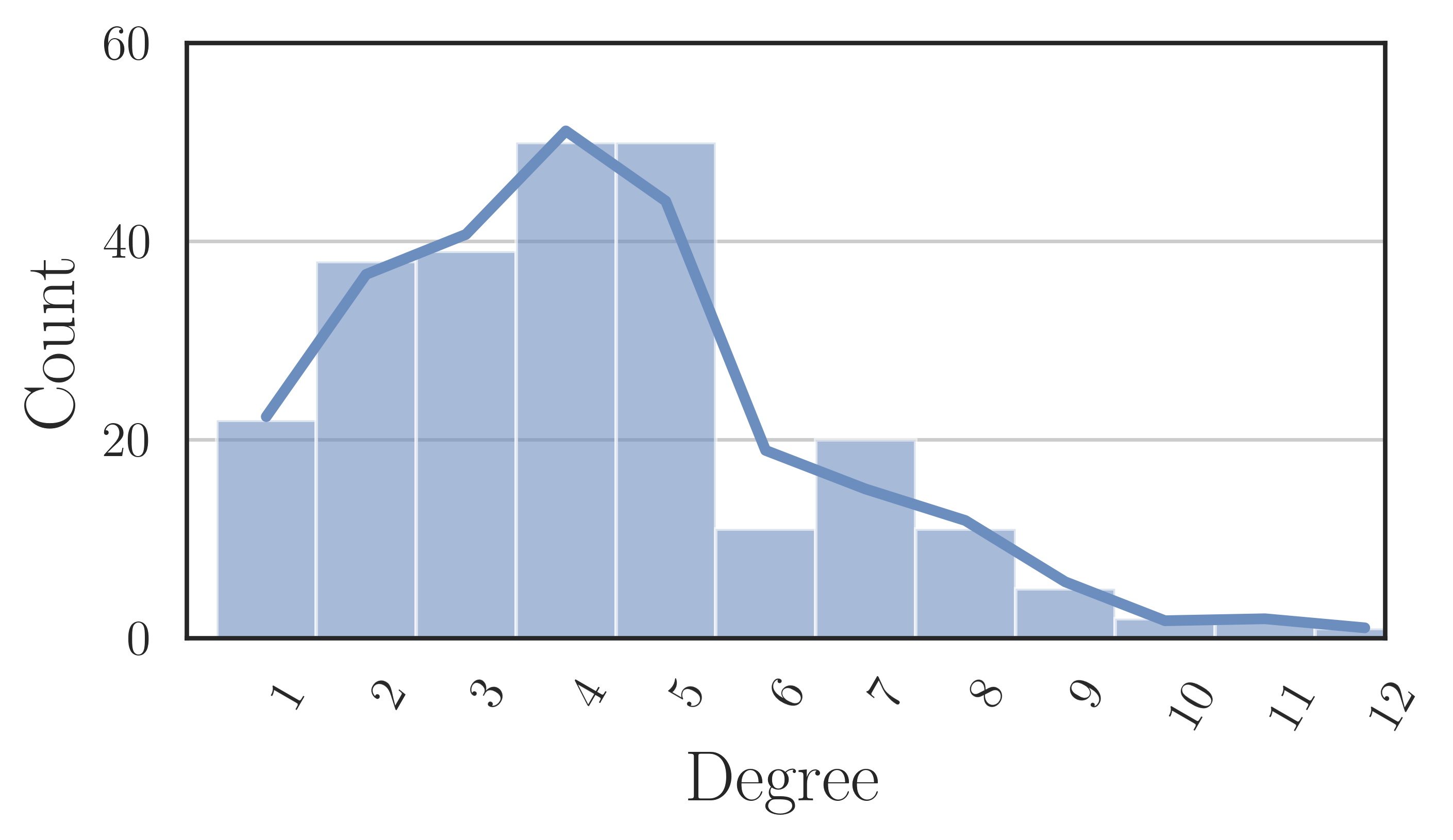}
    \vspace{-8pt}
  \end{subfigure}
  \begin{subfigure}[b]{0.25\linewidth}
    \centering
    \includegraphics[width=\textwidth]{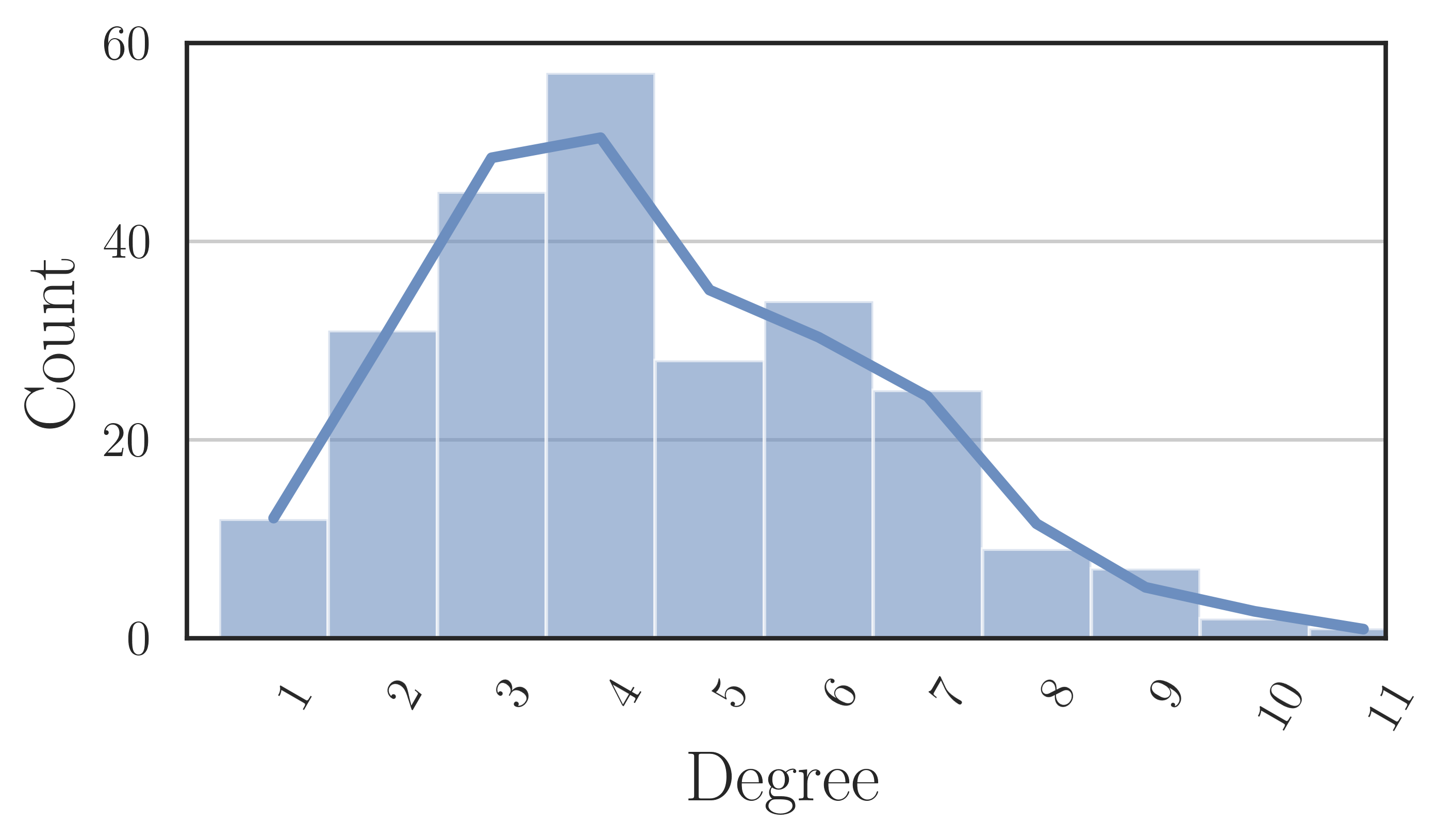}
    \vspace{-8pt}
  \end{subfigure}
  \begin{subfigure}[b]{0.25\linewidth}
    \centering
    \includegraphics[width=\textwidth]{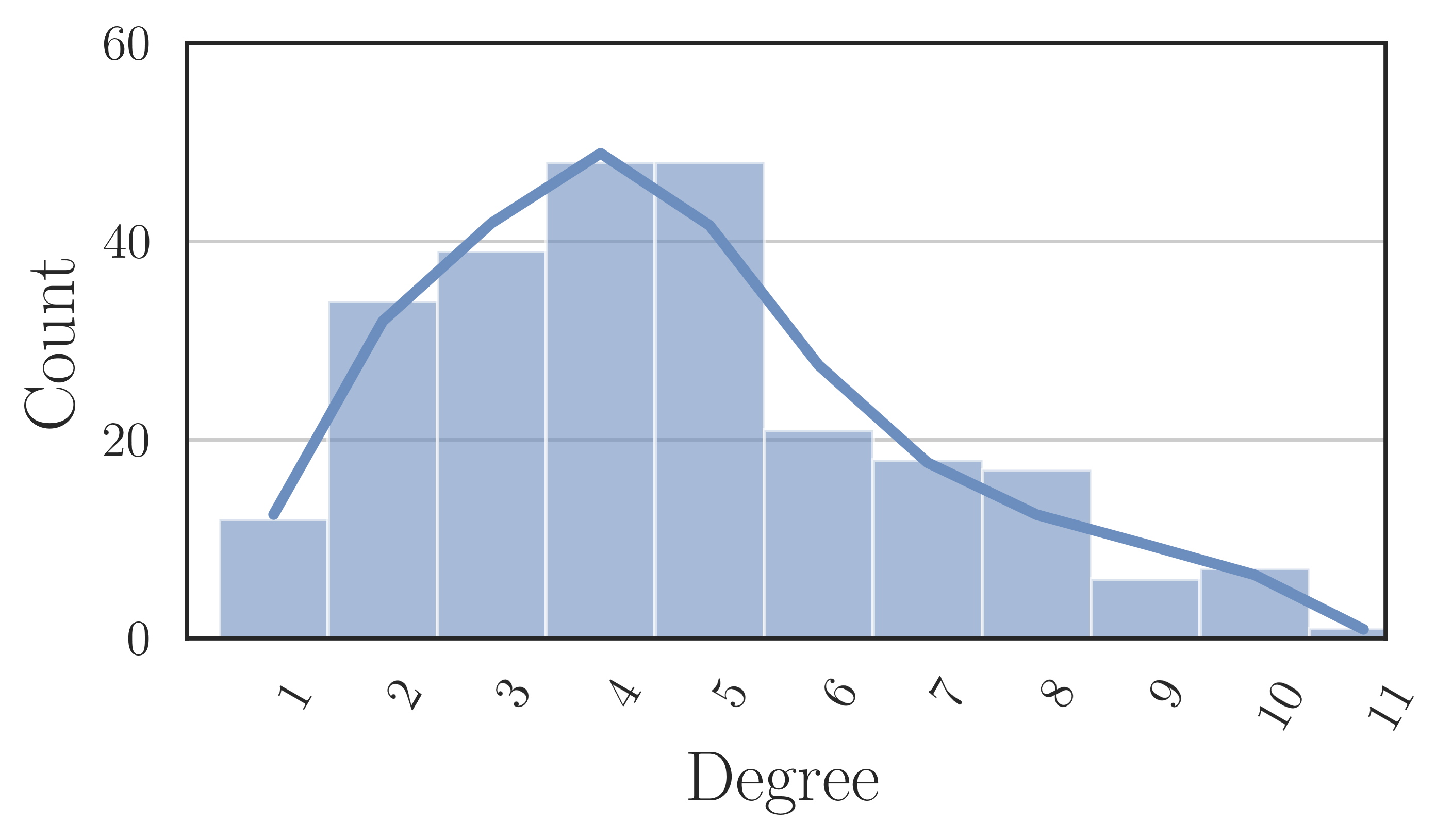}
    \vspace{-8pt}
  \end{subfigure}

    \begin{subfigure}[b]{0.25\linewidth}
    \centering
    \includegraphics[width=\textwidth]{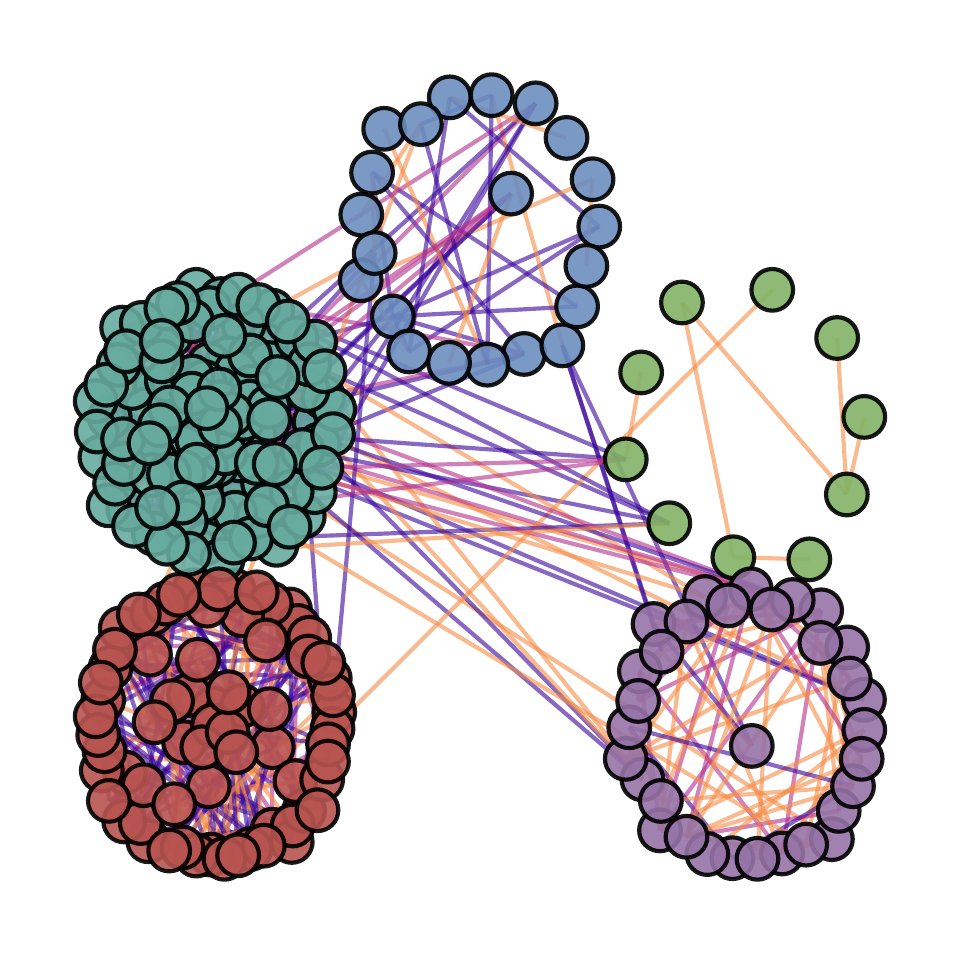}
    \vspace{-8pt}
    \caption{Epoch 100, $\%_{p} = 55$ $h=0.96$}
  \end{subfigure}
  \begin{subfigure}[b]{0.25\linewidth}
    \centering
    \includegraphics[width=\textwidth]{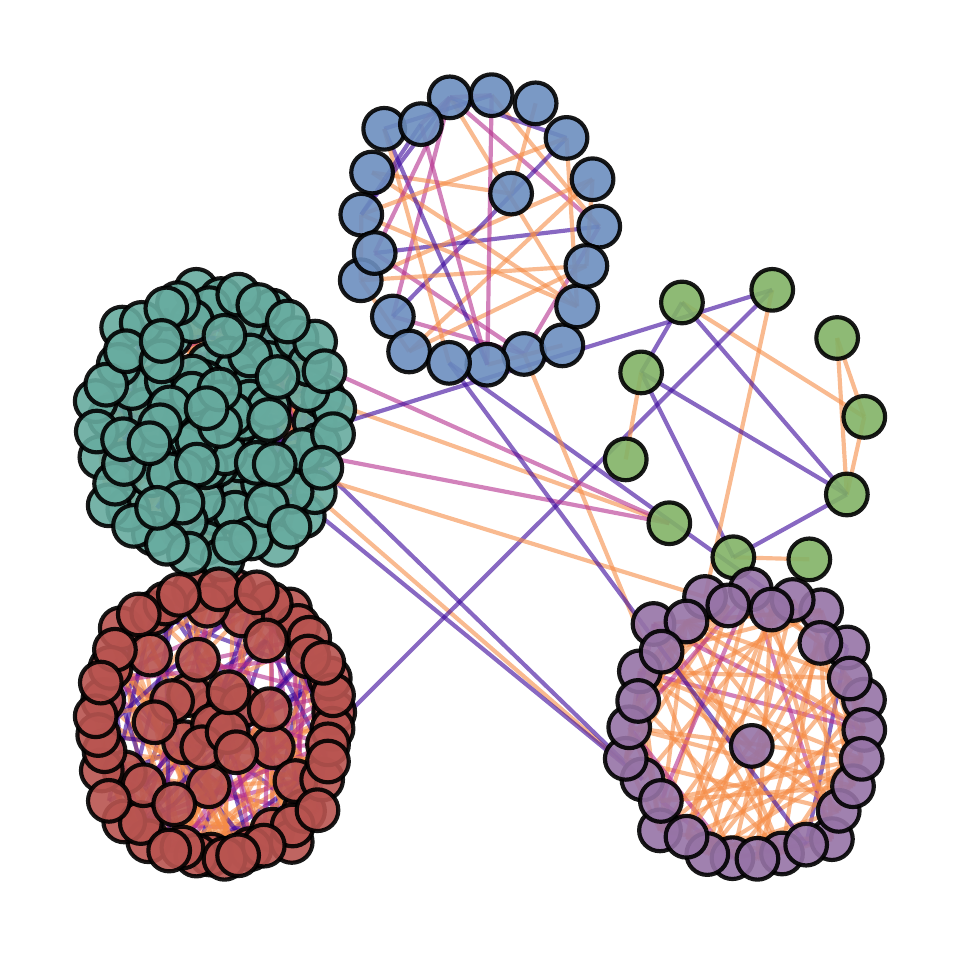}
    \vspace{-8pt}
    \caption{Epoch 150, $\%_{p} = 36 $, $h=0.98$}
  \end{subfigure}
  \begin{subfigure}[b]{0.25\linewidth}
    \centering
    \includegraphics[width=\textwidth]{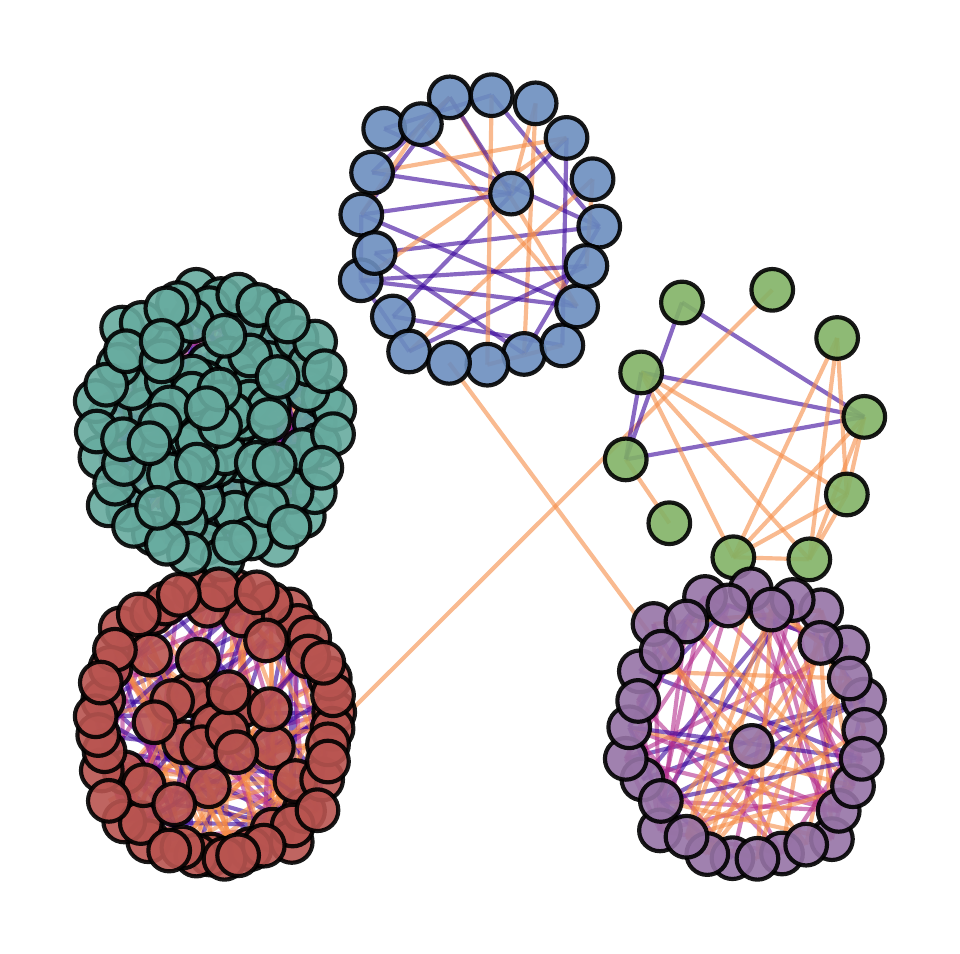}
    \vspace{-8pt}
    \caption{Epoch 200, $\%_{p} = 18 $, $h=0.99$}
  \end{subfigure}

    \caption{Evolution of the latent complex for the Wisconsin dataset, along with homophily level and nodes degree distribution. Edges in \textcolor{edge_col}{orange}, triangles in \textcolor{triangle_col}{lilac}, squares in \textcolor{square_col}{purple} ($K_{max} = 4$)}
    \label{fig:plots_wis}
\end{figure*}

In this appendix we show additional plots of the inferred latent cell complexes for the Wisconsin dataset. Due to the higher number of nodes of the other heterophilic datasets, we believe that these visualizations are less helpful than those in the main text. The plots showed in Figure \ref{fig:plots_wis}, as the ones showed in Figure \ref{fig:plots}, are obtained by training the architecture without providing the input graph and further show the ability of the DCM to learn a latent complexes with high level of homophily. Moreover, in Tables \ref{tab:abl_homlev_homo} and \ref{tab:abl_homlev_hetero} we show the average homophily level of the learned graphs for all the tested datasets, both when the input graph is provided or not. It is worth noticing that the homophily levels of the learned graphs are always greater than the input graph, even on the homophilic datasets.

\section{Model Architecture}
\label{ap:arch}
In this appendix, we present a detailed description of the architectures employed to obtain the results in Table \ref{tab:res-homo} and Table \ref{tab:res-hetero}. To ensure uniformity  and show the performance gain without intensive ad-hoc hyperparameters tuning (as the ones performed for the DGM \cite{kazi2022dgm} and the DGM-M \cite{borde2023latent}), we maintained a constant configuration for the number of layers, hidden dimensions, activation functions, $K_{max}$ (4), (pseudo-)similarity functions (minus the euclidean distances among embeddings), dropout rates (0.5), and learning rates (0.01) across all datasets. The architecture details are shown in Tables \ref{tab:arch} and \ref{tab:dcm_arch}. We conducted training for a total of 200 epochs for the homophilic datasets, with the exception of the physics dataset, which underwent 100 epochs like the heterophilic datasets. Our experiments were performed using a single NVIDIA RTX A6000 with 48 GB of GDDR6 memory. As mentioned in Section $\ref{sec:experiments}$, in every experiment we employ Graph Convolutional Neural Networks (GCNs) and Cell Complex Convolutional Neural Network (CCCNs) as specific MP architectures at node and edge levels, respectively. We give a brief description of GCNs and CCCNs in the following. 

\subsection{Graph and Cell Complex Convolutional Neural Networks}
\textbf{Graph Convolutional Neural Networks} (GCNs) are one of the most famous and simple GNNs architectures. In GCNs, the ouput features of a node are computed as a weighted sum of linearly transformed features of its neighboring nodes. Therefore, the MP round in \eqref{int_node} is implemented as:

\begin{equation}
\mathbf{x}_{0,int}(i) = \gamma_0\left(\sum_{j \in \mathcal{N}_u(\sigma_i^0)} a_{i,j} \mathbf{x}_{0,in}(j)\mathbf{W} \right),
\end{equation}

where $\mathbf{W}$ are learnable parameters and the (normalized) weights $a_{i,j}$ are set as $a_{i,j}=\frac{1}{\sqrt{d_j d_i}}$, with  \(d_i\) and \(d_j\) being the degrees of node \(i\) and node \(j\), respectively \cite{kipf2016semi}. 

\textbf{Cell Complex Convolutional Neural Networks} (CCCNs) generalize GCNs to cell complexes using the adjacencies introduced in Definition 1. In our case, the output features of an edge are computed as a weighted sum of linearly transformed features of its neighboring edges, over the upper and lower adjacencies. Therefore, in this paper we implement the MP round in \eqref{out_edge} as:

\begin{equation}
\mathbf{x}_{1,out}(i) = \gamma_1\left(\sum_{j \in \mathcal{N}_u(\sigma_i^1)} a_{u,i,j} \mathbf{x}_{1,int}(j)\mathbf{W}_u + \sum_{j \in \mathcal{N}_d(\sigma_i^1)} a_{d,i,j} \mathbf{x}_{1,int}(j)\mathbf{W}_d + \mathbf{x}_{1,int}(j)\mathbf{W} \right),
\end{equation}
where $\mathbf{W}_u$, $\mathbf{W}_d$, and $\mathbf{W}$ are learnable parameters. The weights are normalized with the same approach of GCNs, with upper (for the $a_{u,i,j}$s) and lower (for the $a_{d,i,j}$s) degrees. The skip connection $\mathbf{x}_{1,int}(j)\mathbf{W}$ is as usual beneficial, and in the case of CCCNs it has a further theroretical justification in terms of Hodge Decomposition and signal filtering, see \cite{roddenberry2022cellsp, sardellitti2022cell} for further details. 

\begin{table}[h!]\vspace{-.3cm}
  \caption{Homophily level of the latent graph on homophilic graph node classification benchmarks.}
  \label{tab:abl_homlev_homo}
  \centering
\scalebox{.86}{\begin{tabular}{lc ccccc}
    \toprule   & Cora & CiteSeer & PubMed & Physics & CS \\

    \hline

    w/o graph & 0.93 & 0.96 & 0.84 & 0.98 & 0.99\\
    \hline

    w graph & 0.89 & 0.78 & 0.82 & 0.96 & 0.92\\
    \hline
    input & 0.81 & 0.74 & 0.80 & 0.93 & 0.80\\
    \bottomrule
    
    \end{tabular}}
    
\end{table}

\begin{table}[h!]\vspace{-.5cm}
  \caption{Homophily level of the latent graph on heterophilic graph node classification benchmarks.}
  \label{tab:abl_homlev_hetero}
  \centering
\scalebox{.86}{\begin{tabular}{lc cccc}
    \toprule   & Texas & Wisconsin & Squirrel & Chameleon  \\

    \hline

    w/o graph & 0.99 & 0.99 & 0.3 &  0.64\\
    \hline

    w graph & 0.80 & 0.70 & 0.29 & 0.62\\
    \hline
    input & 0.11 & 0.21 & 0.22  & 0.23\\
    \bottomrule
    
    \end{tabular}}
    
\end{table}

\begin{table}[h!]\vspace{-.5cm}
  \caption{Model Architecture}
  \label{tab:arch}
  \centering
\begin{tabular}{l l l }
    \toprule
    No. Layer param.  & Activation & LayerType \\

    \hline
    (no. input features, 32) & ReLU & Linear \\    
    \hline
      &   & DCM \\
     \hline 
    (32, 32) & ReLU & Graph Conv \\
    (32, 32) & ReLU & Cell Conv \\
    \hline
    (64, no. classes) & Softmax & Linear \\
    \bottomrule
    
    \end{tabular}
    
\end{table}
\begin{table}[h!] \vspace{-.5cm}
  \caption{DCM Architecture}
  \label{tab:dcm_arch}
  \centering
\begin{tabular}{lccc}
\toprule & & DCM* & DCM \\
\hline No. Layer param. & Activation & \multicolumn{2}{c}{ Layer type } \\
\hline (32, 32) & ReLU & Linear & Graph Conv \\
(32, 32) & ReLU & Linear & Graph Conv \\
(32, 32) & None & Linear & Graph Conv \\
\bottomrule
\end{tabular}
\end{table}
\end{appendix}

\end{document}